\documentclass[review]{elsarticle}

\usepackage{lineno}
\usepackage{url}
\usepackage{xcolor}
\usepackage{multirow}
\usepackage{float}
\usepackage{graphicx}
\usepackage{subcaption}
\usepackage{threeparttable}
\usepackage{eurosym}
\usepackage{float}
\usepackage{framed,multirow}
\usepackage{pifont}
\usepackage{amsmath}
\modulolinenumbers[5]

\journal{Accident Analysis \& Prevention}

\begin{document}

\begin{frontmatter}


\title{Detecting motorcycle helmet use with deep learning}






\author[1]{Felix Wilhelm Siebert\corref{cor1}} 
\cortext[cor1]{Corresponding author:
  Tel.: +49-(0)30-314-229-67;  
  }
  \ead{felix.siebert@tu-berlin.de}
\author[2]{Hanhe Lin}
\address[1]{Department of Psychology and Ergonomics, Technische Universit{\"a}t Berlin, Marchstra{\ss}e 12, 10587 Berlin, Germany}
\address[2]{Department of Computer and Information Science, Universit{\"a}t Konstanz, Universit{\"a}tsstra{\ss}e 10, 78464 Konstanz, Germany}
\begin{abstract}
The continuous motorization of traffic has led to a sustained increase in the global number of road related fatalities and injuries. To counter this, governments are focusing on enforcing safe and law-abiding behavior in traffic. However, especially in developing countries where the motorcycle is the main form of transportation, there is a lack of comprehensive data on the safety-critical behavioral metric of motorcycle helmet use. This lack of data prohibits targeted enforcement and education campaigns which are crucial for injury prevention. Hence, we have developed an algorithm for the automated registration of motorcycle helmet usage from video data, using a deep learning approach. Based on 91,000 annotated frames of video data, collected at multiple observation sites in 7 cities across the country of Myanmar, we trained our algorithm to detect active motorcycles, the number and position of riders on the motorcycle, as well as their helmet use. An analysis of the algorithm's accuracy on an annotated test data set, and a comparison to available human-registered helmet use data reveals a high accuracy of our approach. Our algorithm registers motorcycle helmet use rates with an accuracy of -4.4\% and +2.1\% in comparison to a human observer, with minimal training for individual observation sites. Without observation site specific training, the accuracy of helmet use detection decreases slightly, depending on a number of factors. Our approach can be implemented in existing roadside traffic surveillance infrastructure and can facilitate targeted data-driven injury prevention campaigns with real-time speed. Implications of the proposed method, as well as measures that can further improve detection accuracy are discussed.
\end{abstract}
\date{} 
\begin{keyword}
Deep learning \sep Helmet use detection \sep Motorcycle \sep Road safety \sep Injury prevention
\end{keyword}

\end{frontmatter}


\section{Introduction}
\label{sec:intro}
Using a motorcycle helmet can decrease the probability of fatal injuries of motorcycle riders in road traffic crashes by 42\% \citep{liu2004helmets} which is why governments worldwide have enacted laws that make helmet use mandatory. Despite this, compliance with motorcycle helmet laws is often low, especially in developing countries \citep{bachani2013trends, bachani2017helmet, siebert2019patterns}. 
To efficiently conduct targeted helmet use campaigns, it is essential for governments to collect detailed data on the level of compliance with helmet laws. 
However, 40\% of countries in the world do not have an estimate of this crucial road safety metric \citep{world2015global}. And even if data is available, helmet use observations are frequently limited in sample size and regional scope \citep{fong2015smallsample, ledesma2015academic}, draw from data of relatively short time frames \citep{karuppanagounder2016academic, xuequn2011time}, or are only singularly collected in the scope of academic research \citep{siebert2019patterns, oxley2018academic}. The main reason for this lack of comprehensive continuous data lies in the prevailing way of helmet use data collection, which utilizes direct observation of motorcycle helmet use in traffic by human observers. This direct observation during road-side surveys is resource intensive, as it is highly time-consuming and costly \citep{eby2011naturalistic}. And while the use of video cameras allows \textit{indirect} observation, alleviating the time pressure of \textit{direct} observation, the classification of helmet use through human observers limits the amount of data that can be processed.

In light of this, there is an increasing demand to develop a reliable and timely efficient intelligent system for detecting helmet use of motorcycle riders that does not rely on a human observer. A promising method for achieving this automated detection of motorcycle helmet use is machine learning. Machine learning has been applied to a number of road safety related detection tasks, and has achieved high accuracy for the general detection of pedestrians, bicyclists, motorcyclists and cars \cite{dalal2005histograms}. While first implementations of automated motorcycle helmet use detection have been promising, they have not been developed to their full potential. Current approaches rely on data sets that are limited in the overall number of riders observed, are trained on a small number of observation sites, or do not detect the rider position on the motorcycle \citep{dahiya2016automatic,vishnu2017detection}. In this paper a deep learning based automated helmet use detection is proposed that relies on a comprehensive dataset with large variance in the number of riders observed, drawing from multiple observation sites at varying times of day. 

Recent successful deep learning based applications of computer vision, e.g. in image classification \citep{he2016deep,simonyan2014very,szegedy2016rethinking}, object detection \citep{lin2018focal,he2017mask}, and activity recognition \citep{pigou2018beyond,donahue2015long} have heavily relied on large-scale datasets, similar to the one used in this study. Hence, the next section of this paper will focus on the generation of the underlying dataset and its annotation, to facilitate potential data collection in other countries with a similar methodology. This is followed by a section on algorithm training. In the subsequent sections of this paper, the algorithm performance is analyzed through comparison with an annotated test data set and with results from an earlier observational study on helmet use in Myanmar, conducted by human observers \citep{siebert2019patterns}.


\section{Dataset creation and annotation}

\subsection{Data collection and preprocessing} \label{Data collection and preprocessing}

\begin{table}[!hb]
\centering
\caption{254 hours of source video were available from 13 observation sites in 7 cities across Myanmar, from which 1,000 video clips (10 seconds / 100 frames each) were sampled for further annotation.}
\resizebox{0.8\textwidth}{!}{
\begin{tabular}{l r l r r} \hline
& &  & \multicolumn{1}{c}{Duration}&\multicolumn{1}{c}{Sampled} \\
City&Population &Site ID&  (hours)& clips \\\hline \hline
\multirow{3}{*}{Bago}&\multirow{3}{*}{254,424}&\texttt{Bago\_highway}&9 & 35\\
&&\texttt{Bago\_rural} & 17 & 67\\
&&\texttt{Bago\_urban} & 16 & 63\\\hline
\multirow{2}{*}{Mandalay}& \multirow{2}{*}{1,225,546}&\texttt{Mandalay\_1}&58&228 \\
&&\texttt{Mandalay\_2}&48& 190\\\hline
\multirow{2}{*}{NayPyiTaw}& \multirow{2}{*}{333,506}&\texttt{Naypyitaw\_1}&13&51 \\
& &\texttt{Naypyitaw\_2}&11& 43\\ \hline
\multirow{2}{*}{Nyaung-U}&\multirow{2}{*}{48,528} &\texttt{NyaungU\_rural}&21&83 \\
&&\texttt{NyaungU\_urban}& 17&67\\\hline
Pakokku& 90,842 &\texttt{Pakokku\_urban} &19&75 \\\hline
\multirow{2}{*}{Pathein}&\multirow{2}{*}{169,773} &\texttt{Pathein\_rural} &3&12 \\ 
& &\texttt{Pathein\_urban} &12& 47\\\hline
Yangon&4,728,524 &\texttt{Yangon\_II} &10&39\\ \hline\hline
&& & 254&1,000\\ 
\end{tabular}
}
\label{tb:dataforanno}
\end{table}

Myanmar was chosen as the basis for the collection of the source material for the development of the algorithm, since its road user aggregate and rapid motorization are highly representative of developing countries in the world \citep{world2017powered} and video recordings of traffic were available from an earlier study \citep{siebert2019patterns}. Motorcyclists comprise more than 80\% of road users in Myanmar \citep{world2015global}, and the number of motorcycles has been increasing rapidly in recent years \citep{Wegman2017road}.  

\begin{figure} [!ht]
    \centering
    \begin{minipage}[b]{0.32\textwidth}\centering
        \includegraphics[width=0.98\textwidth]{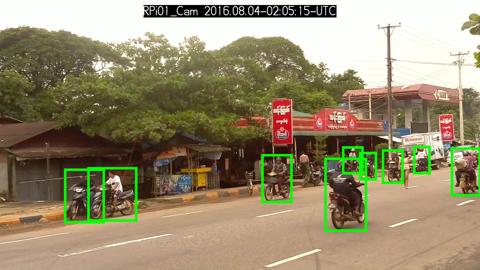}
    \centerline{(a) \texttt{Bago\_highway}}
    \end{minipage}
    \begin{minipage}[b]{0.32\textwidth}\centering
        \includegraphics[width=0.98\textwidth]{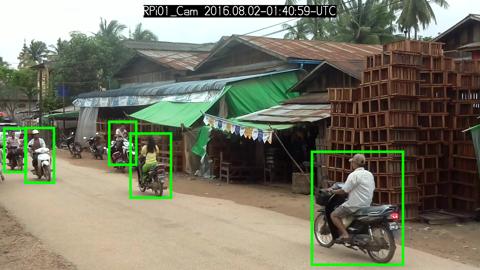}
        \centerline{(b) \texttt{Bago\_rural}}
    \end{minipage}
    \begin{minipage}[b]{0.32\textwidth}\centering
        \includegraphics[width=0.98\textwidth]{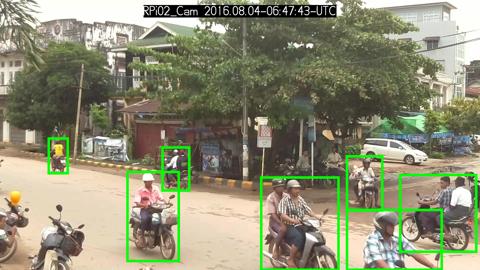}
        \centerline{(c) \texttt{Bago\_urban}}
    \end{minipage}
    \begin{minipage}[b]{0.32\textwidth}\centering
        \includegraphics[width=0.98\textwidth]{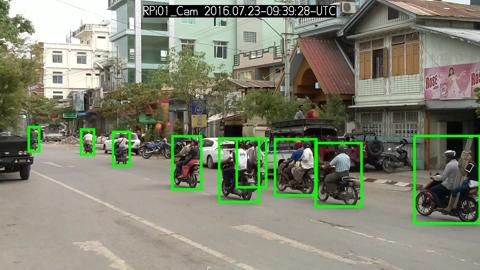}
        \centerline{(d) \texttt{Mandalay\_1}}
    \end{minipage}
    \begin{minipage}[b]{0.32\textwidth}\centering
        \includegraphics[width=0.98\textwidth]{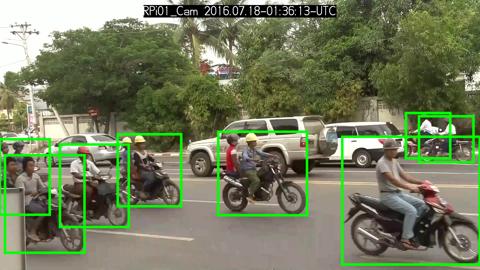}
        \centerline{(e) \texttt{Mandalay\_2}}
    \end{minipage}
    \begin{minipage}[b]{0.32\textwidth}\centering
        \includegraphics[width=0.98\textwidth]{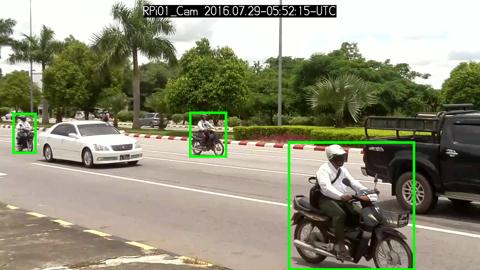}
        \centerline{(f) \texttt{Naypyitaw\_1}}
    \end{minipage}
    \begin{minipage}[b]{0.32\textwidth}\centering
        \includegraphics[width=0.98\textwidth]{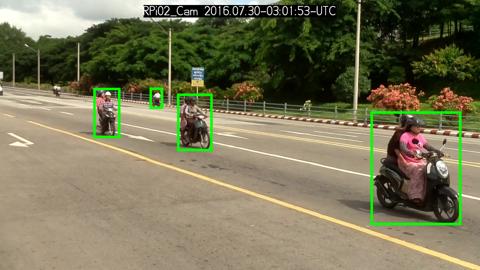}
        \centerline{(g) \texttt{Naypyitaw\_2}}
    \end{minipage}
    \begin{minipage}[b]{0.32\textwidth}\centering
        \includegraphics[width=0.98\textwidth]{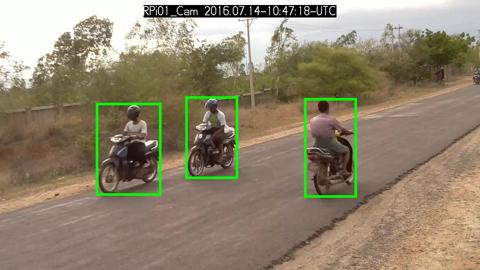}
        \centerline{(h) \texttt{NyaungU\_rural}}
    \end{minipage}
    \begin{minipage}[b]{0.32\textwidth}\centering
        \includegraphics[width=0.98\textwidth]{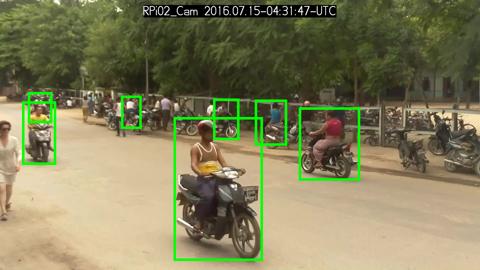}
        \centerline{(i) \texttt{NyaungU\_urban}}
    \end{minipage}
    \begin{minipage}[b]{0.32\textwidth}\centering
        \includegraphics[width=0.98\textwidth]{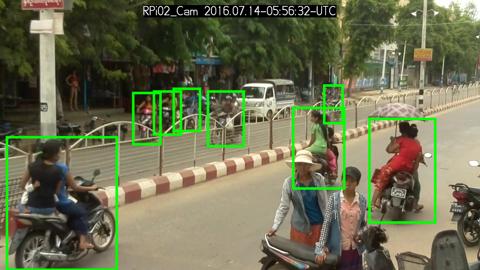}
        \centerline{(j) \texttt{Pakokku\_urban}}
    \end{minipage}
    \begin{minipage}[b]{0.32\textwidth}
    	\centering
        \includegraphics[width=0.98\textwidth]{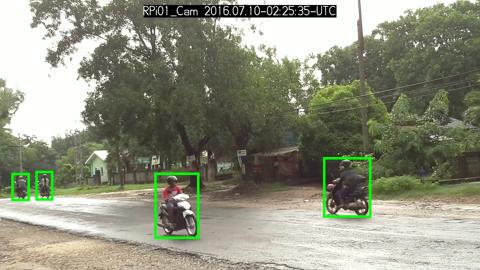}
        \centerline{(k) \texttt{Pathein\_rural}}
    \end{minipage}
    \begin{minipage}[b]{0.32\textwidth}
    	\centering
        \includegraphics[width=0.98\textwidth]{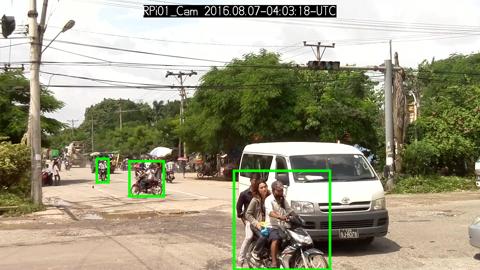}
        \centerline{(l) \texttt{Yangon\_II}}
    \end{minipage}
    \caption{Still frames in 12 observation sites of 7 cities throughout Myanmar, where green rectangles correspond to annotations.}\label{fig:obsite}
\end{figure}

Throughout Myanmar, traffic was filmed with two video-cameras with a resolution of $1920\times1080$ pixels and a frame rate of 10 frames per second. Within seven cities, cameras were placed at multiple observation sites at approximately 2.5 m height and traffic was recorded for two consecutive days from approximately 6 am to 6:30 pm (Table~\ref{tb:dataforanno}).
As the city of Mandalay has the highest number of motorcyclists in Myanmar the two cameras were installed for 7 days here. Yangon, the largest city of Myanmar, has an active ban on motorcycle in the city center, hence, one camera was placed in the suburbs here. Due to technical problems with the cameras and problems in reaching the selected observation sites, the number of hours recorded was not the same for each observation site. After the removal of blurred videos due to cloudy weather or rain, 254 hours of video data were available as the source material for this study. Video data was divided into 10 second video clips (100 frames each), which formed the basis for training, validating, and testing the algorithm in this study. The duration of video data available at each observation site is shown in Table~\ref{tb:dataforanno}. The observation sites represent a highly diverse data set, including multilane high traffic density road environments (e.g. Mandalay) as well as more rural environments (e.g. Pathein). Still frames of observation sites are presented in Figure \ref{fig:obsite}.

\subsection{Sampling video clips}

Since there were insufficient resources to annotate all 254 hours of recorded traffic, 1,000 videos clips were sampled which were most representative of the source material. After segmenting the source material into non-overlapping video clips of 10 seconds length (100 frames), we applied the object detection algorithm YOLO9000 \citep{redmon2017yolo9000} with the pre-trained weights to detect the number of motorcycles in each frame, extracting those clips with the highest number of motorcycles in them.
Multiple clips were sampled from each observation site, in proportion to the available videodata from each site. The resulting distribution of the 1,000 sampled video clips is presented in Table~\ref{tb:dataforanno}. The observation site \texttt{Pathein\_urban} (47 video clips) was retroactively excluded from analysis due to heavy fogging on the camera which was not detected during the initial screening of the video data (Section \ref{Data collection and preprocessing}). In addition, 43 video clips were excluded since they did not contain active motorcycles, as the YOLO9000 algorithm \citep{redmon2017yolo9000} had identified parked motorcycles.


\subsection{Annotation}
\label{sec:annot}
\begin{figure*} [!ht]
    \centering
    \begin{subfigure}[b]{0.45\textwidth}
        \includegraphics[width=0.95\textwidth]{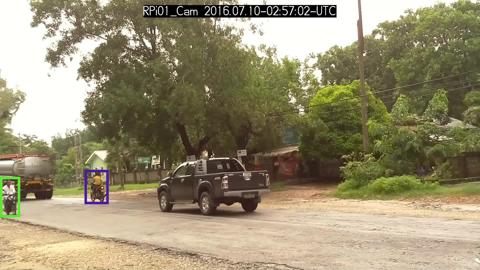}
        \caption{}
    \end{subfigure}
    \begin{subfigure}[b]{0.45\textwidth}
        \includegraphics[width=0.95\textwidth]{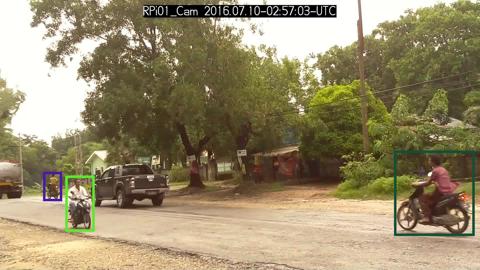}
        \caption{}
    \end{subfigure}    
    \begin{subfigure}[b]{0.45\textwidth}
        \includegraphics[width=0.95\textwidth]{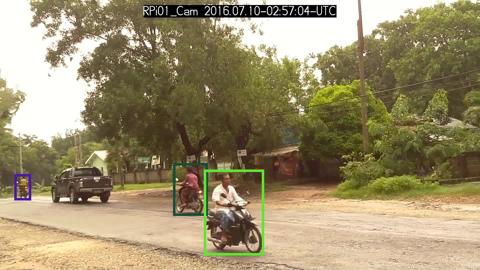}
        \caption{}
    \end{subfigure}
    \begin{subfigure}[b]{0.45\textwidth}
        \includegraphics[width=0.95\textwidth]
        {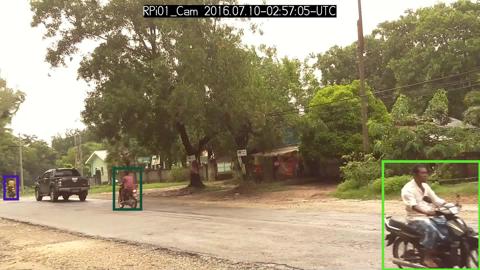}
        \caption{}
    \end{subfigure}
    \caption{An example of motorcycle annotation. An individual motorcycle (marked in light green rectangles) appears on the left side of the frame and disappears on the lower right side of the frame.}
    \label{fig:track_annot}
\end{figure*}

Videodata was annotated by first drawing a rectangular box around an individual motorcycle and its riders (so called \textit{bounding box}), before entering information on the number of riders, their helmet use and position. All bounding boxes containing an individual motorcycle throughout a number of frames form the \textit{motorcycle track}, i.e. an individual motorcycle will appear in multiple frames, but will only have one \textit{motorcycle track}. To facilitate the annotation of the videos, we tested and compared the three image and video annotation tools BeaverDam \citep{Shen:EECS-2016-193}, LabelMe \citep{russell2008labelme}, and VATIC \citep{vondrick2013efficiently}. We chose BeaverDam for data annotation, since it allows frame-by-frame labeling in videos, is easy to install, and has superior usability. Annotation was conducted by two freelance workers. An example of the annotation of an individual motorcycle through multiple frames (\textit{motorcycle track}) is presented in Fig.~\ref{fig:track_annot}.

\begin{figure}[!htb]
    \centering
    \includegraphics[width=0.30\textwidth]{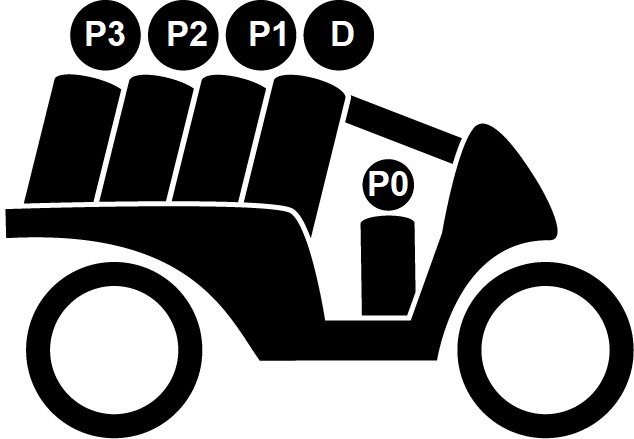}
    \includegraphics[width=0.30\textwidth]{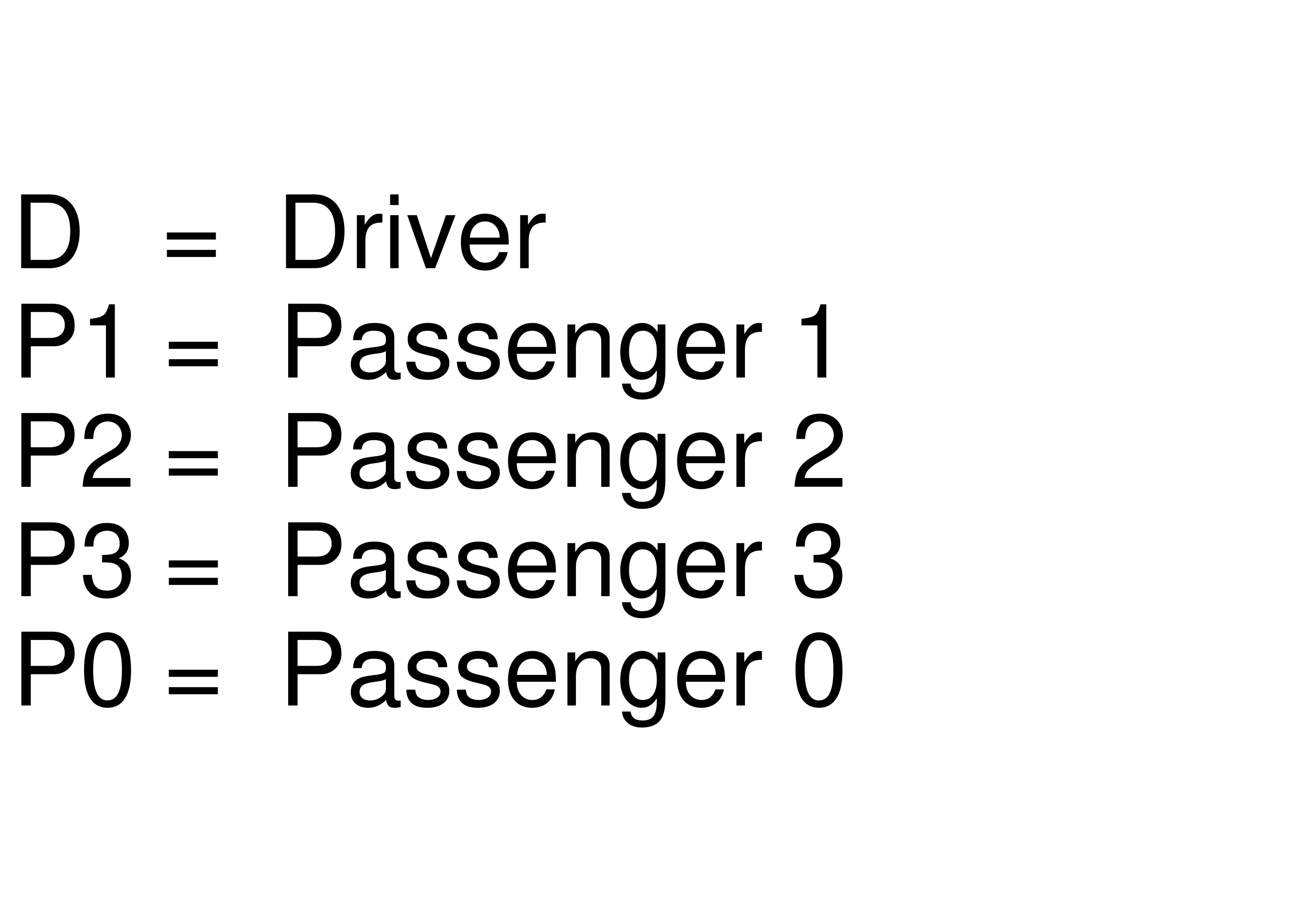}
    \centerline{(I)}
    
    \begin{minipage}[b]{0.185\textwidth}
        \includegraphics[width=\textwidth]{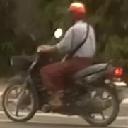}
        \centerline{(a)}
    \end{minipage}
    \begin{minipage}[b]{0.185\textwidth}
        \includegraphics[width=\textwidth]{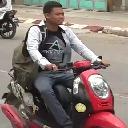}
        \centerline{(b)}
    \end{minipage}    
    \begin{minipage}[b]{0.185\textwidth}
        \includegraphics[width=\textwidth]{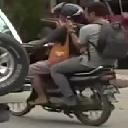}
        \centerline{(c)}
    \end{minipage}
    \begin{minipage}[b]{0.185\textwidth}
        \includegraphics[width=\textwidth]{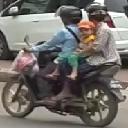}
        \centerline{(d)}
    \end{minipage}
    \begin{minipage}[b]{0.185\textwidth}
        \includegraphics[width=\textwidth]
        {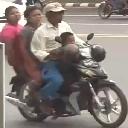}
        \centerline{(e)}
    \end{minipage}
    \centerline{(II)}
    \caption{Structure (I) and examples (II) of helmet use encoding: (a) \texttt{DHelmet}, (b) \texttt{DNoHelmet}, (c) \texttt{DHelmetP1NoHelmet}, (d) \texttt{DHelmetP1NoHelmetP2Helmet}, and (e) \texttt{DNoHelmetP0NoHelmetP1NoHelmetP2NoHelmet}.}
    \label{fig:helencode}
\end{figure}

For each bounding box, workers encoded the number of riders (1 to 5), their helmet use (yes/no) and position (Driver (D), Passenger (P0-3); Fig.~\ref{fig:helencode}). Examples of rider encoding are displayed in Fig.~\ref{fig:helencode}.


\begin{table}[!hbt]
\footnotesize
\centering
\caption{910 annotated video clips were randomly split into training, validation and test sets according to individual observation site, with a split ratio of 70\%, 10\%, and 20\%.}
\begin{tabular}{ l| r| r| r ||r }\hline
  Site ID       & Training set & Validation set & Test set & \textbf{Overall} \\ \hline
  \texttt{Bago\_highway} & 24 & 4 & 7 & 35  \\
  \texttt{Bago\_rural} & 41 & 6 & 11& 58  \\
  \texttt{Bago\_urban} &44 & 6 &13 & 63 \\
  \texttt{Mandalay\_1} &159 & 23 &45& 227 \\
  \texttt{Mandalay\_2} &111  &16 & 31& 158 \\
  \texttt{Naypyitaw\_1}&36&5 &10& 51 \\
  \texttt{Naypyitaw\_2}&30 &4 &9& 43 \\
  \texttt{NyaungU\_rural}&57 &8 &17& 82 \\
  \texttt{NyaungU\_urban}&47&7&13& 67 \\
  \texttt{Pakokku\_urban}& 52 & 8&15& 75 \\
  \texttt{Pathein\_rural}& 8 &1 &3& 12 \\
  \texttt{Yangon\_II}& 27 & 4 & 8&39 \\ \hline \hline
  \textbf{Overall}& 636 & 92 & 182 &910 \\
\end{tabular}
\label{tb:dataset}
\end{table}

\subsection{Composition of annotated data}

The 910 annotated video clips were randomly divided into three non-overlapping subsets: a training set (70\%), a validation set (10\%), and a test set (20\%) (Table \ref{tb:dataset}). Data on the number of annotated motorcycles in all 910 video clips can be found in Table \ref{tb:instancestat_and_accuracy}. Overall, 10,180 motorcycle tracks (i.e. individual motorcycles) were annotated. As each individual motorcycle appears in multiple frames, there are 339,784 annotated motorcycles on a frame level, i.e. there are 339,784 bounding boxes containing motorcycles in the dataset. All motorcycles were encoded in classes, depending on the position and helmet use of the riders. This resulted in 36 classes, shown in Table \ref{tb:instancestat_and_accuracy}. The number of motorcycles per class was imbalanced and ranged from only 12 observed frames (e.g., for motorcycles with 5 riders with no rider wearing a helmet) to 140,886 observed frames (one driver wearing a helmet). Some classes were not observed in the annotated video clips, e.g., there was no motorcycle with 4 riders all wearing a helmet.

\section{Helmet use detection algorithm}

\subsection{Method} \label{Method_ML}
\begin{figure}
	\centering
	\includegraphics[width=1.0\textwidth]{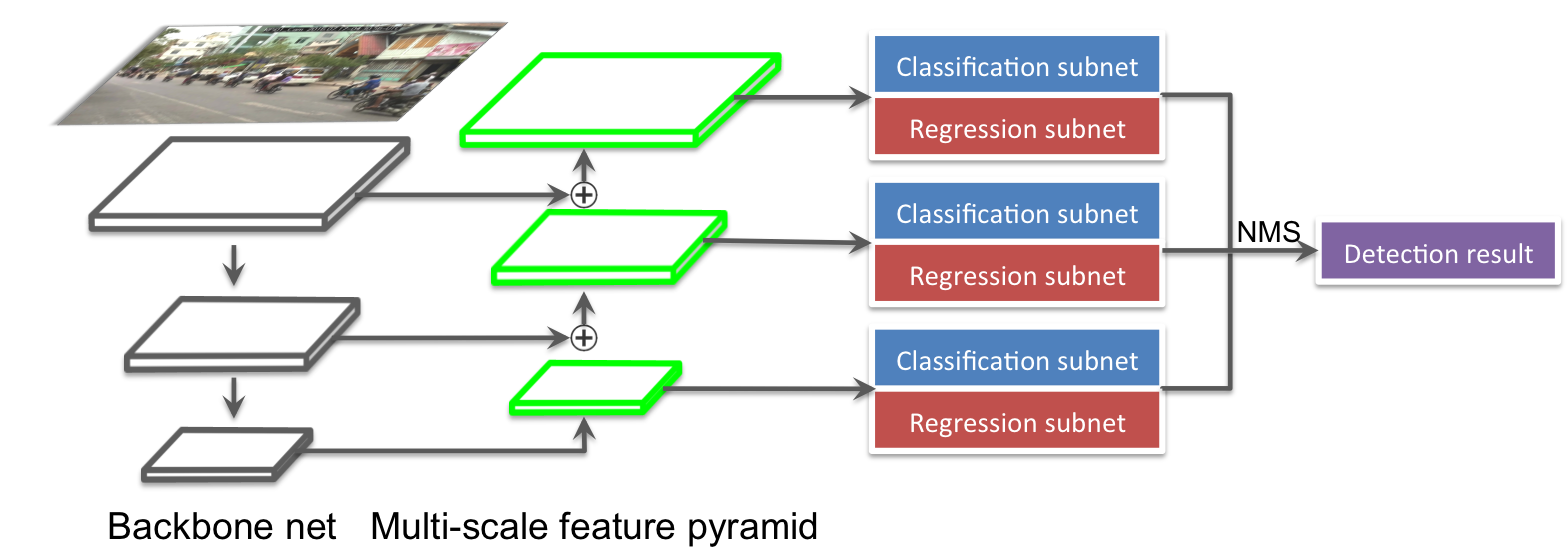}
	\caption{The framework of RetinaNet \cite{lin2018focal}. A given image is first fed into a backbone convolutional neural network to generate a multi-scale feature map, from which a multi-scale feature pyramid is generated. In each level of the pyramid, there are two subnetworks. One is for regression from anchor boxes to ground-truth object boxes, the other is for classifying anchor boxes. After non-maximum suppression (NMS) across the multi-scale feature pyramid, RetinaNet arrives at the detection results.}\label{fig:retinanet}
\end{figure}

After the creation of the dataset was finished, we applied a state-of-the-art object detection algorithm to the annotated data, to facilitate motorcycle helmet use detection on a frame-level. In this process, data from the \textit{training set} is used to train the object detection algorithm. In the process of training, the \textit{validation set} is used to find the best generalizing model, before the algorithm's accuracy in predicting helmet use is tested on data that the algorithm has not seen before, the so-called \textit{test set}. The composition of the three sets is presented in Table \ref{tb:dataset}. 
Generally, the state-of-the-art object detection algorithms can be divided into two types: two-stage and single-stage approaches. The two-stage approaches first identify a number of potential locations within an image, where objects could be located. In a second step, an object classifier (using a convolutional neural network) is used to identify objects a these locations. While two-stage approaches such as Fast R-CNN \cite{girshick2015fast}, achieve a higher accuracy than single-stage approaches, they are very time-consuming. In contrast, single-stage approaches simultaneously conduct object location and object identification. Single stage approaches like YOLO \cite{redmon2017yolo9000} and RetinaNet \cite{lin2018focal} therefore are much faster than two-stage approaches, although there is a small trade-off in accuracy. In this paper, we used RetinaNet \cite{lin2018focal} for our helmet use detection task. While it is a single-stage approach, it uses a multi-scale feature pyramid and focal loss to address the general limitation of one-stage detectors in accuracy. Figure~\ref{fig:retinanet} illustrates the framework of RetinaNet. 

\subsection{Training} \label{Training_ML}

Since the task of detecting motorcycle riders' helmet use is a classic object detection task, we fine-tuned RetinaNet instead of training it from scratch. I.e. we use a RetinaNet model\footnote{https://github.com/fizyr/keras-retinanet} which is already trained for general object detection and fine tune it to specifically detect motorcycles, riders, and helmets.

\begin{figure}[!htb]
    \centering
    \begin{subfigure}[b]{0.45\textwidth}
        \includegraphics[width=\textwidth]{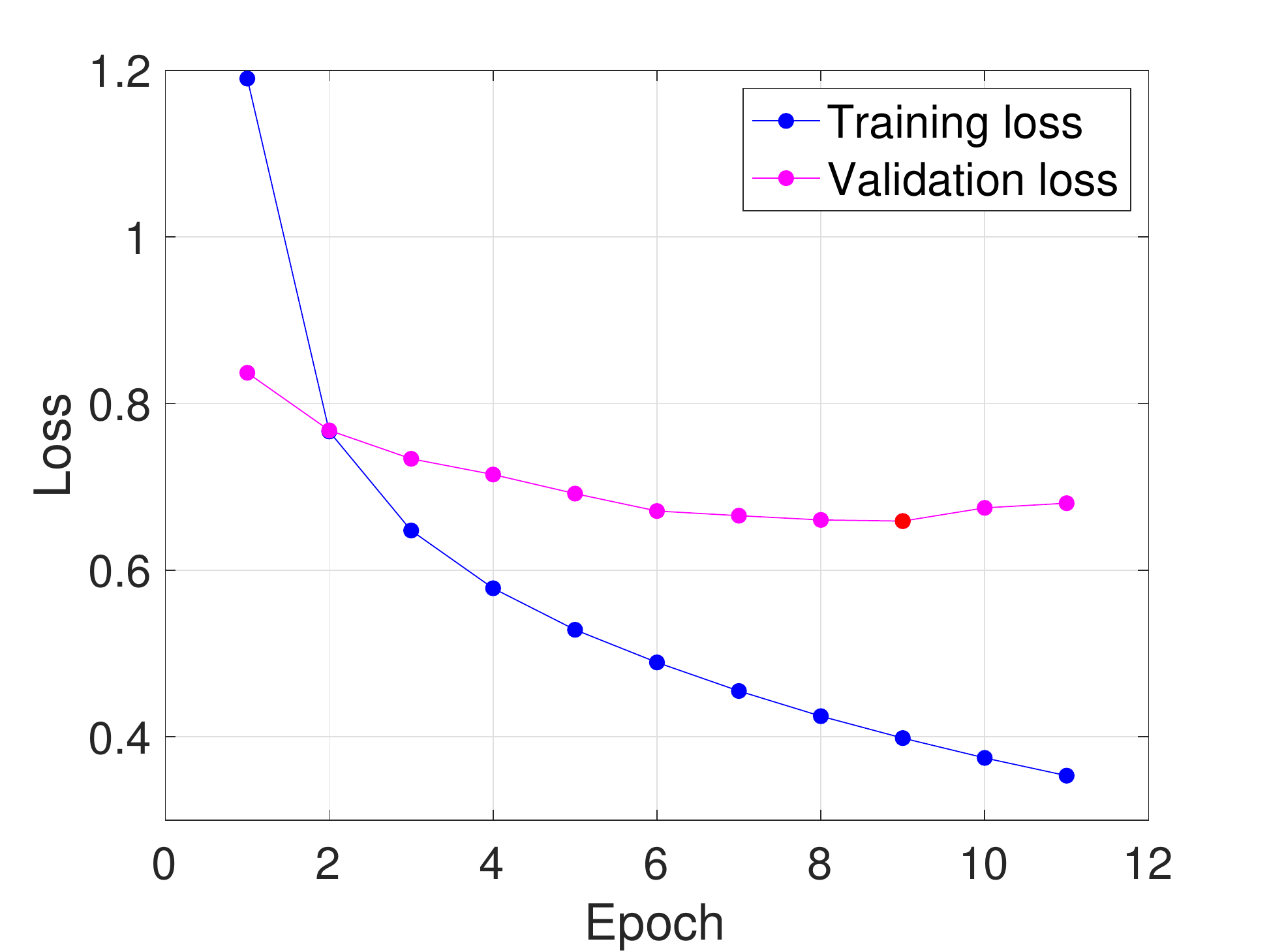}
        \caption{}
    \end{subfigure}
    \begin{subfigure}[b]{0.45\textwidth}
        \includegraphics[width=\textwidth]{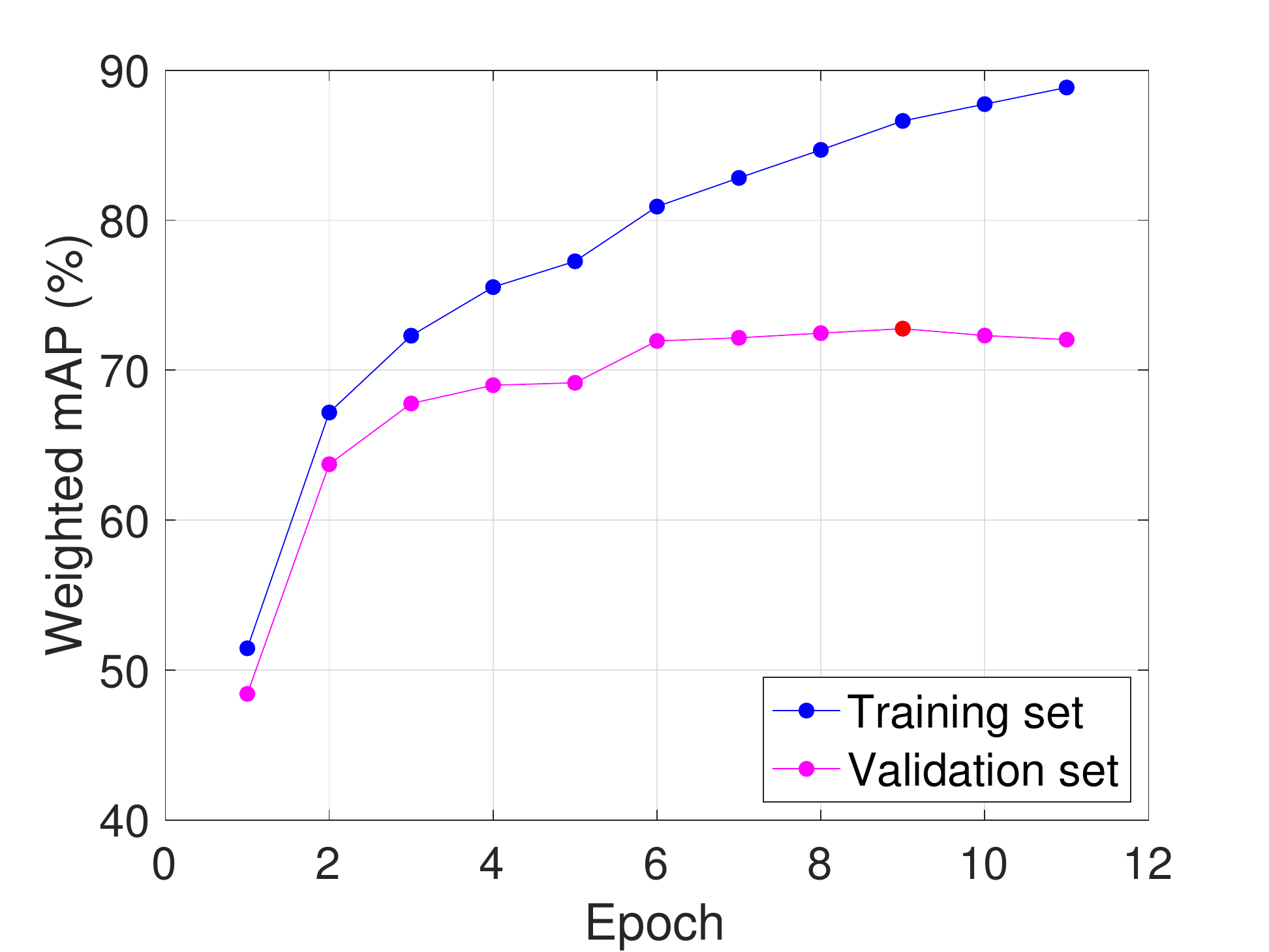}
        \caption{}
    \end{subfigure}
    \caption{The learning process of RetinaNet for helmet use detection. (a) training and validation loss, (b) weighted mAP on training and validation set. The algorithm achieved 72.8\% weighted mAP (red point) on the validation set after 9 epochs.}
    \label{fig:trainProgress}
\end{figure}

In our experiments, we used ResNet50 \citep{he2016deep} as the backbone net, initialized with pre-trained weights from ImageNet \cite{deng2009imagenet}. The backbone net provides the specific architecture for the convolutional neural network. In the learning process, we used the Adam optimizer \cite{kingma2014adam} with a learning rate of $\alpha = 0.00001$ and a batch size of 4 and stopped training when the weighted mean Average Precision (weighted mAP, explained in the following) on the validation set stopped improving with a patience of 2. To assess the accuracy of our algorithm, we use the Average Precision (\textit{AP}) value \cite{everingham2010pascal}. The \textit{AP} integrates multiple variables to produce a measure for the accuracy of an algorithm in an object detection task, including \textit{intersection over union},
\textit{precision}, and \textit{recall}. The \textit{intersection over union} describes the positional relation between algorithm generated and human annotated bounding boxes. Algorithm generated bounding boxes need to overlap with human annotated bounding boxes by at least 50\%, otherwise they are registered as an incorrect detection. The \textit{precision} presents the number of correct detections of all detections made by the algorithm (\(\textit{precision}=\frac{\text{true positives}}{\text{true positives + false positives}}\)). The \textit{recall} variable measures how many of the available correct instances were detected by the algorithm (\(\textit{recall}=\frac{\text{true positives}}{\text{true positives + false negatives}}\)). For a more in-depth explanation of \textit{AP} please see \cite{everingham2010pascal} and \cite{salton1983introduction}.   
Since the number of frames per class was very imbalanced in our dataset (Table \ref{tb:instancestat_and_accuracy}), the final performance for all classes is computed as weighted average of \textit{AP} for each class, defined as:
\begin{equation}
    \text{weighted mAP} = \sum_{i=1}^C w_i \text{AP}_i,
\end{equation}
where weights $w_i$ across all $C$ classes will sum to one, and set to be proportional to the number of instances. 
Fig.~\ref{fig:trainProgress} shows the training loss, validation loss, and weighted \textit{mAP} in the training and validation sets in the learning process. It can be observed that training loss is constantly decreasing, i.e. the prediction error is getting smaller, while the deep model learns useful knowledge for the helmet use detection from the training set. Consequently, the weighted mAP of the training set is constantly increasing. At the same time, the validation loss, i.e. the prediction error on the validation set is getting smaller in the first 9 epochs. Correspondingly, the mAP on the \textit{validation set} is increasing in the first few epochs before it stops to improve after 9 epochs, which means the algorithm starts to overfit on the training set. Therefore, we stopped training and selected the optimal model after 9 epochs, obtaining 72.8\% weighted \textit{mAP} on the validation set.  


Our models were implemented using the Python Keras library with Tensorflow as a backend \citep{chollet2015keras} and ran on two NVIDIA Titan Xp GPUs.




\subsection{Results} \label{Results_ML}

In the following, we report the helmet use detection results of the algorithm on the \textit{test set}, using the optimal model developed on the \textit{validation set} (where it obtained 72.8\% weighted mAP). 

\begin{figure*}[!htb]
    \centering
    \begin{subfigure}[b]{0.42\textwidth}\centering
        \includegraphics[width=1.0\textwidth]{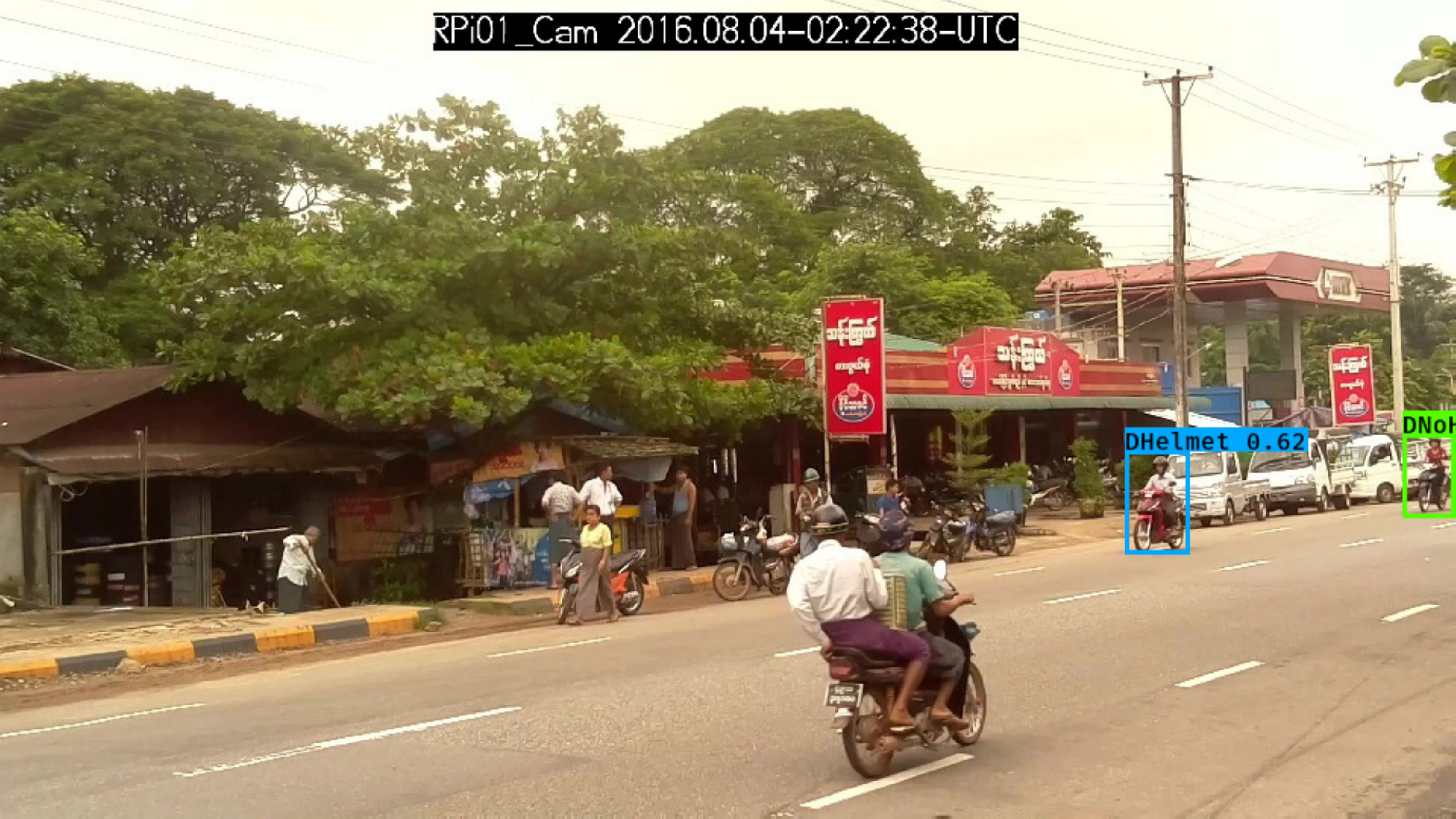}
        \caption{}
    \end{subfigure}
    \begin{subfigure}[b]{0.42\textwidth}\centering
        \includegraphics[width=1.0\textwidth]{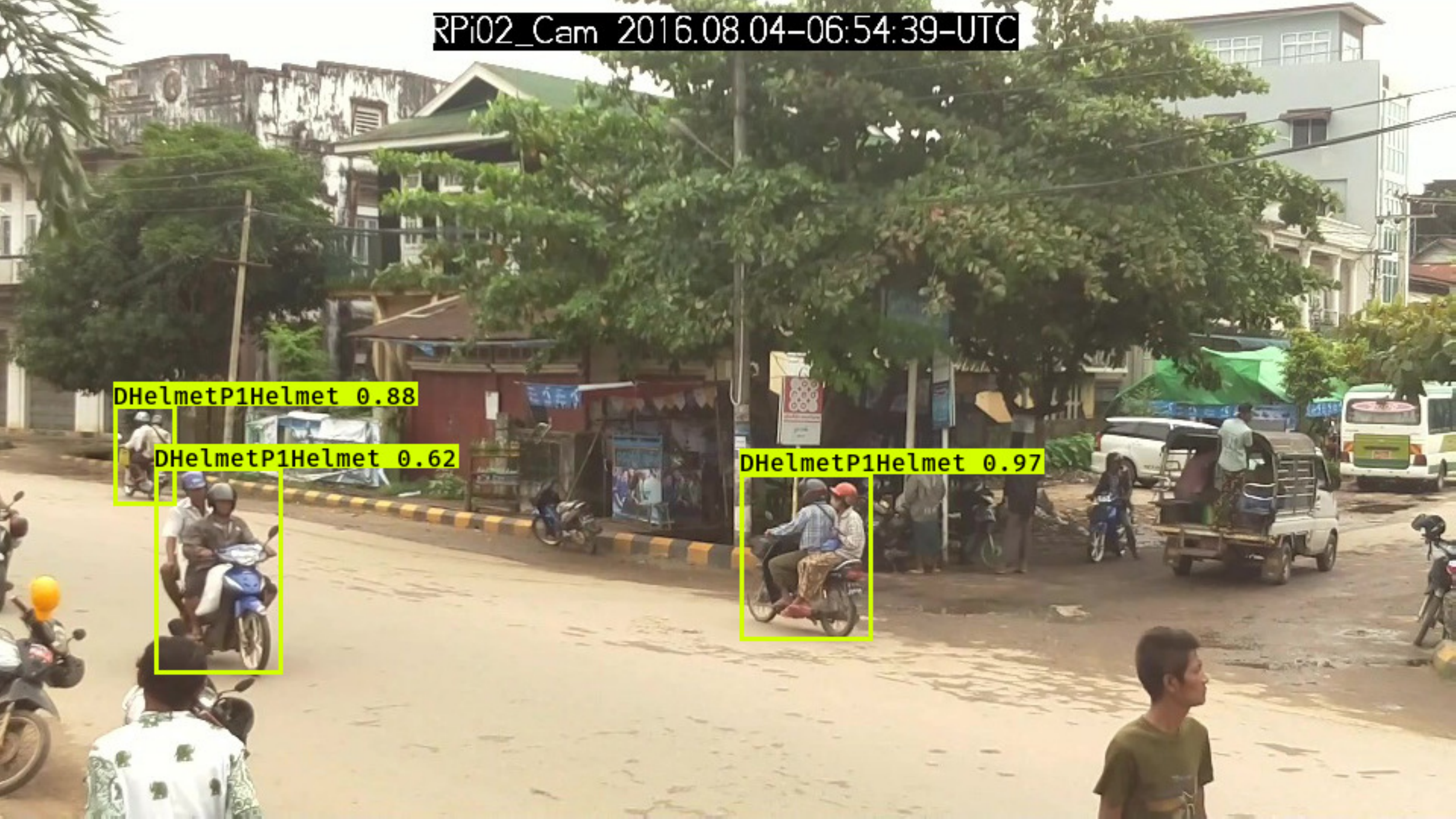}
        \caption{}
    \end{subfigure}
    \hspace{1pt}
    \begin{subfigure}[b]{0.42\textwidth}\centering
        \includegraphics[width=1.0\textwidth]{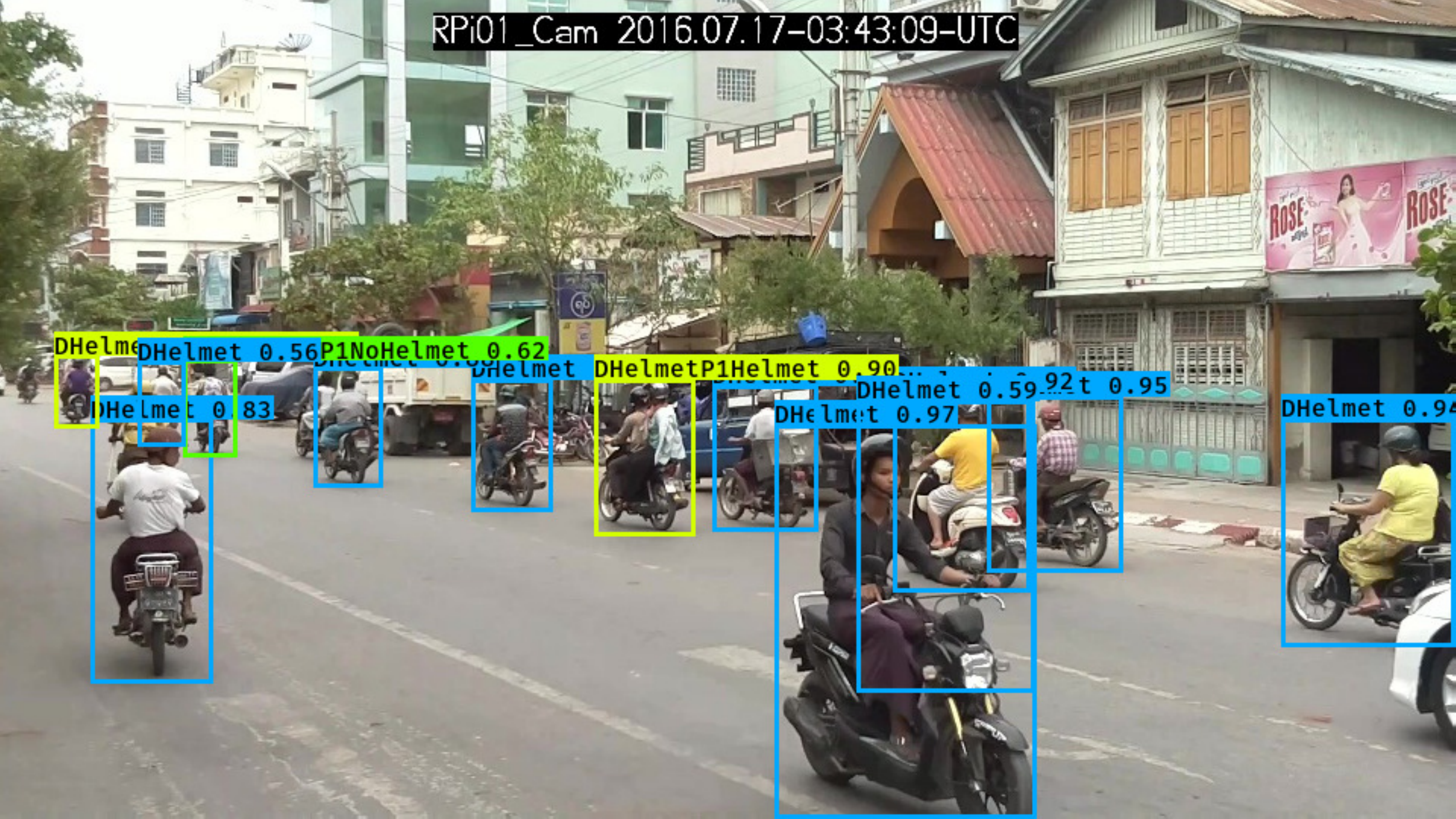}
        \caption{}
    \end{subfigure}
    \begin{subfigure}[b]{0.42\textwidth}\centering
        \includegraphics[width=1.0\textwidth]{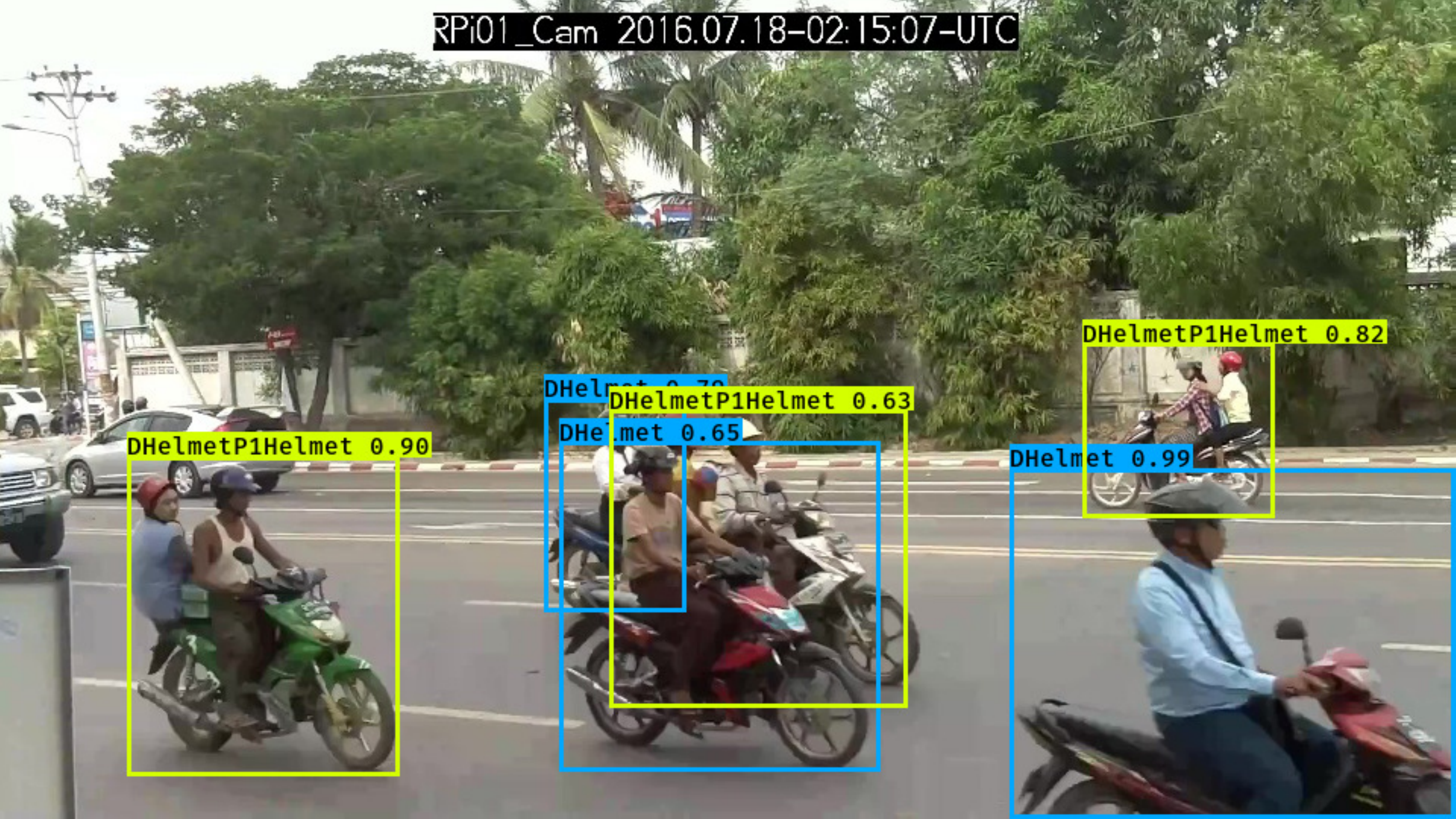}
        \caption{}
    \end{subfigure}
    \hspace{1pt}
    \begin{subfigure}[b]{0.42\textwidth}\centering
        \includegraphics[width=1.0\textwidth]{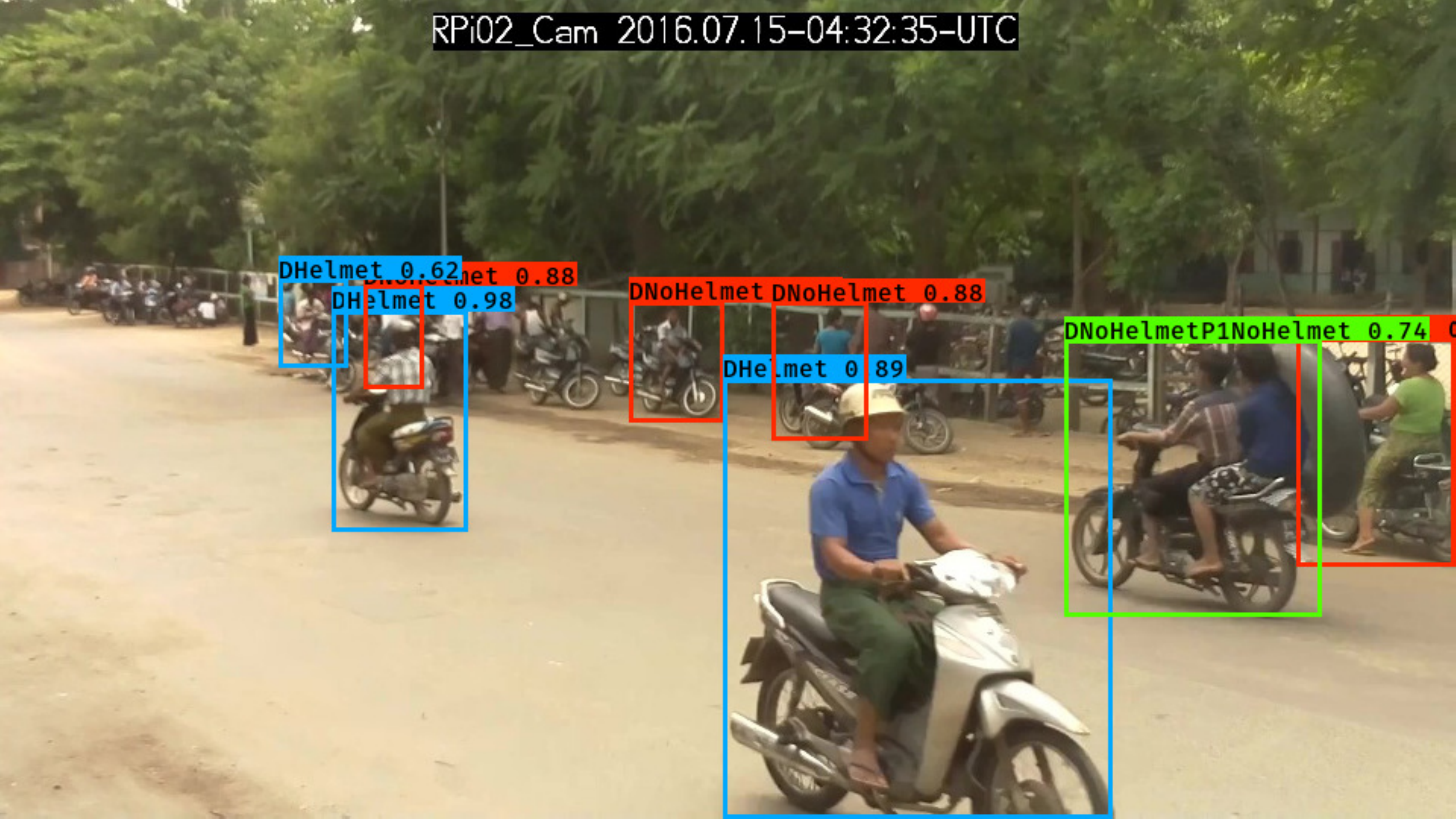}
        \caption{}
    \end{subfigure}
    \begin{subfigure}[b]{0.42\textwidth}\centering
        \includegraphics[width=1.0\textwidth]{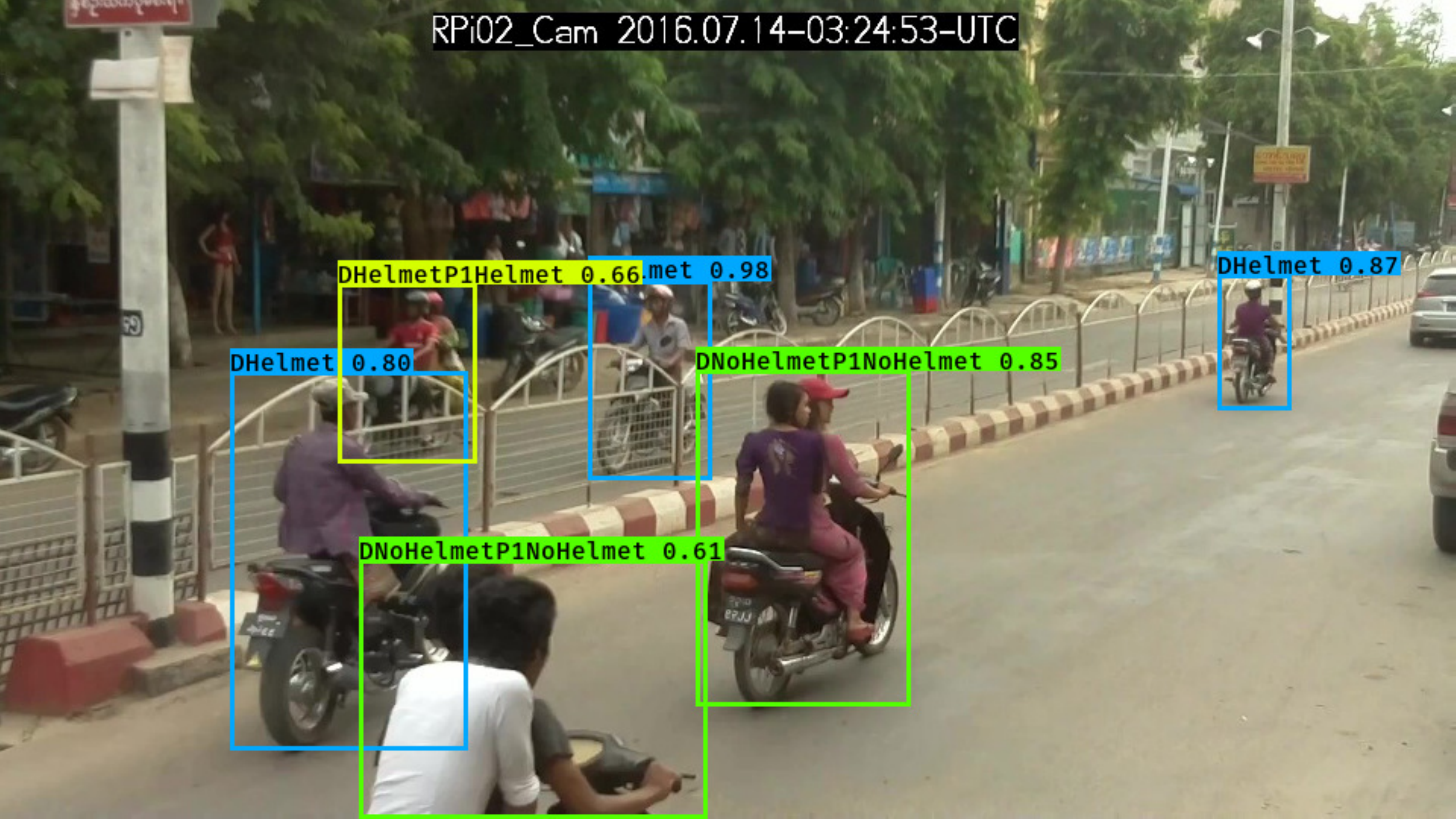}
        \caption{}
    \end{subfigure}
    \hspace{1pt}
    \begin{subfigure}[b]{0.42\textwidth}\centering
        \includegraphics[width=1.0\textwidth]{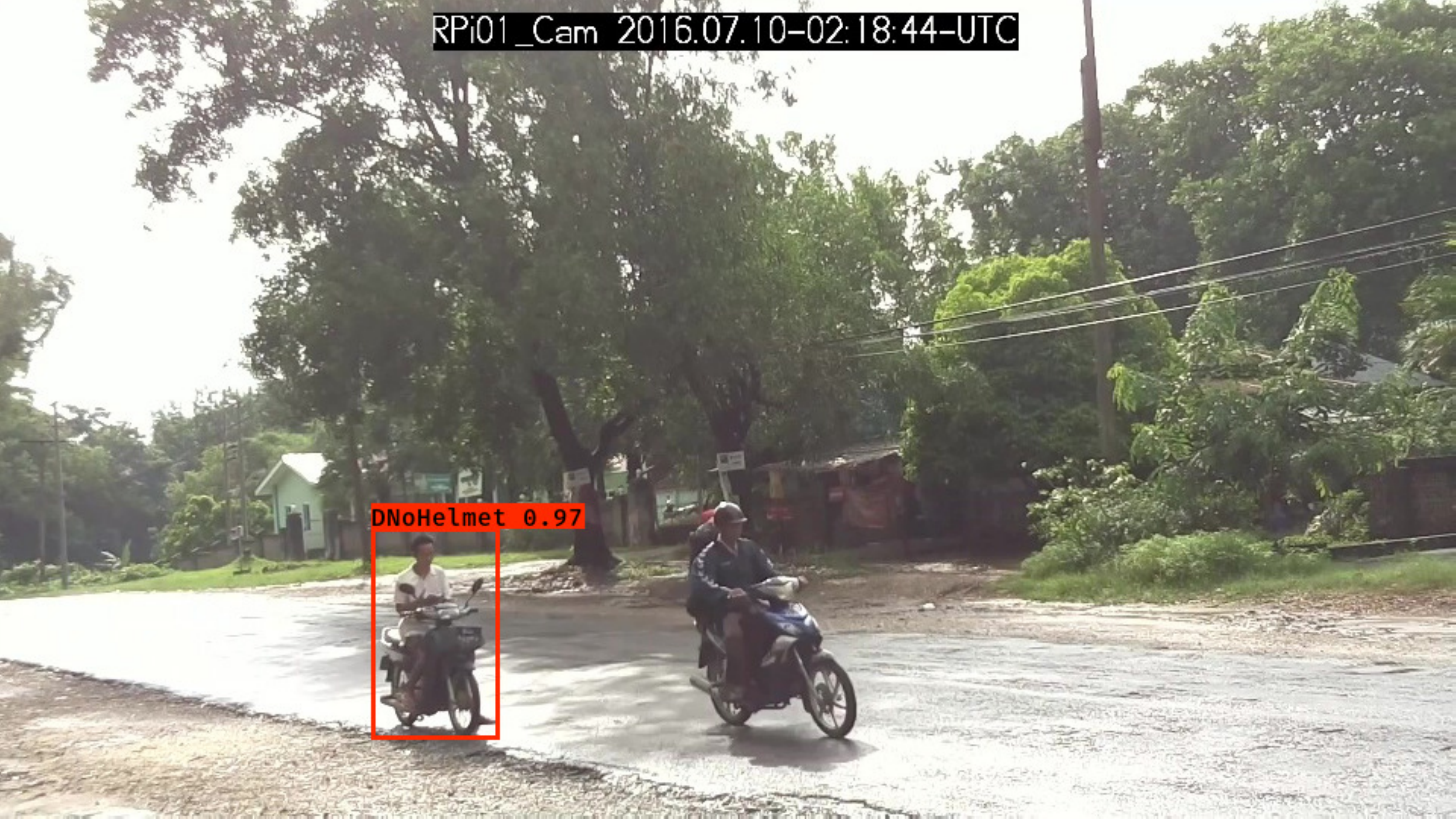}
        \caption{}
    \end{subfigure}
    \begin{subfigure}[b]{0.42\textwidth}\centering
        \includegraphics[width=1.0\textwidth]{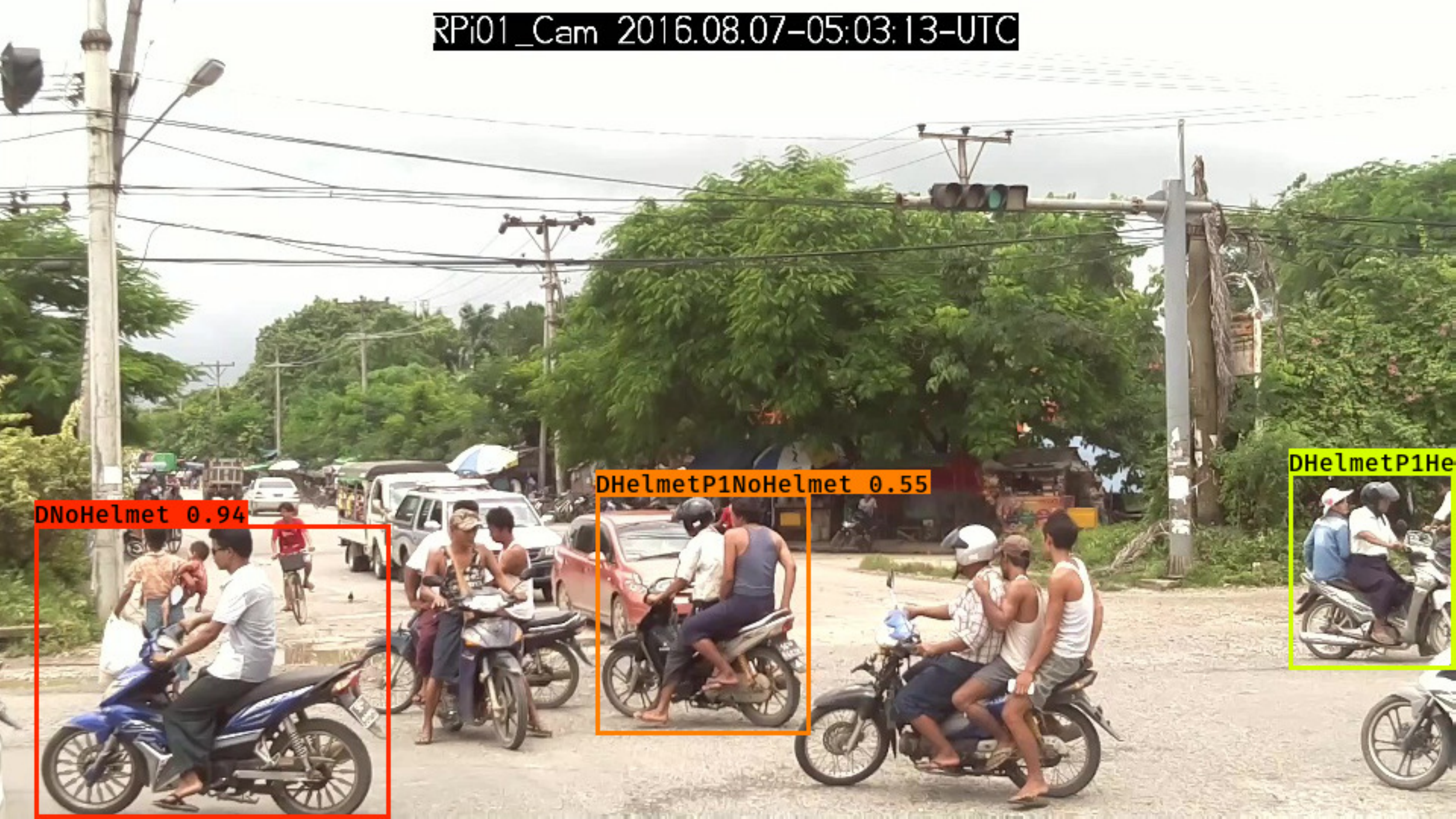}
        \caption{}
    \end{subfigure}
    \caption{Helmet use detection results on sampled frames using RetinaNet. Bounding box colors correspond to different predicted classes.}\label{fig:detectresult}
\end{figure*}

We achieved \textbf{72.3\%} weighted mAP on the test set, with a processing speed of \textbf{14} frames/second. The AP for each class is shown in Table~\ref{tb:instancestat_and_accuracy}. It can be observed that RetinaNet worked well on common classes but not on under-represented classes due to the small number of training instances. Considering only common classes (up to two riders), our trained RetinaNet achieved \textbf{76.4\%} weighted mAP. This is a very good performance considering a lot of factors such as occlusion, camera angle, and diverse observation sites. Detection results on some sample frames are displayed in Fig.~\ref{fig:detectresult}. Due to the imbalanced classes, there are some missing detections, e.g., Fig.~\ref{fig:detectresult} (a), (g) and (h). Example videos, consisting of algorithm annotated frames of the \textit{test set} can be found in the \textit{supplementary material}.

\begin{table}[!htb]

\centering
\caption{Composition of annotated data. 339,784 motorcycles were annotated on a frame level. The last column shows the generalized helmet use detection accuracy (mAP= mean Average Precision).}
\resizebox{0.95\textwidth}{!}{
\begin{threeparttable}
\begin{tabular}{|l|c c c c c | r| r r r r| r|}\hline
&\multicolumn{5}{c|}{Position}&\multicolumn{1}{r|}{Motorcycle}&\multicolumn{4}{c|}{Frame level} &\multicolumn{1}{c|}{Helmet use detection}\\
Class& D  & P1 & P2 & P3 & P0&\multicolumn{1}{r|}{tracks}& Training& Validation& Test & \textbf{Overall} & \multicolumn{1}{r|}{AP (\%)}\\ \hline
1&\ding{51}& -- & -- & -- & -- & 4,406 & 99,029 &14,556 & 27,301 &140,886 & 84.5 \\
2&\ding{51}&\ding{51} & -- & -- & -- & 2,268 & 50.206&7,071 &13,748& 71,025 & 78.5\\
3&\ding{55}& -- & -- & -- & -- & 1,241 & 37,664 &5,936 & 10,796 &54,396 & 75.4\\
4&\ding{55} & \ding{55} & -- & -- & -- &929 &22,723&3,499 &5,736 &31,958 &63.5 \\
5&\ding{51} & \ding{55} & -- & -- & -- &432 &10,729&1,556 &2,314 &14,599 &20.4\\
6&\ding{55} & \ding{55} & \ding{55}& -- & --&211 & 5,290 &377 &1,050&6,717 &28.0\\
7&\ding{55}& \ding{51} & -- & -- & -- &129 &2,853&335 &511&3699  & 11.6\\
8&\ding{51} & \ding{55} & \ding{51} & -- &  --&125&2,456&639 &420 &3,515 &8.6\\
9&\ding{51}& \ding{55}& \ding{55} & -- & -- &75 &1,909&269 &442 &2,620 &8.9\\
10&\ding{55}& \ding{55} & -- & -- & \ding{55}&55 &1,215&113 &514 &1,842 &3.5\\
11&\ding{51} & \ding{51}  & -- & -- & \ding{55} &49 &677&115 &466 &1,258 &9.9\\
12&\ding{55}  & \ding{55} & \ding{55}  & --& \ding{55} &35 &471&208 &277 &956 &18.7\\
13&\ding{51} & -- & -- & -- &\ding{55} &34 & 588 &78&369 &1035 &1.6\\
14&\ding{55} & -- & -- & -- & \ding{55} &28 &701& 95&183 &979 &0.3\\
15&\ding{51} &  \ding{55} & -- & -- &  \ding{55}&24 &600&76 &0 &676 &--\\
16&\ding{51} & \ding{51} & \ding{51} & --&  -- &23 &492&13 &75 &580 &5.1\\
17&\ding{51} & \ding{51} & -- & -- & \ding{51} &22 &446&18 &146 &610 &4.2\\
18&\ding{51} & \ding{55}  & \ding{51} & --& \ding{55} &22&410& 81&24 &515 &1.6\\
19&\ding{51} & -- & -- & -- &\ding{51} &12&352&0 &0&352 &--\\
20&\ding{51} & \ding{55}  & \ding{55} & -- &  \ding{55} &11 &225&0 &27 &252 &0.3\\
21&\ding{55} & \ding{55} & \ding{51} & -- & -- &9 &123&93 &0 &216 &--\\
22&\ding{55} &  \ding{55}  & \ding{55} & \ding{55} & --&6 &334&28 &0&362 &--\\
23&\ding{51}& \ding{51} & \ding{55} & -- &  -- &6 &146&0 &0 &146 &--\\
24&\ding{51}& \ding{55}& \ding{55}  & \ding{51}& --  &5 &42&15 &0 &57 &--\\
25&\ding{51} & \ding{55} & \ding{55}  & \ding{55}& -- &4 &50&0 &70&120 &0.4\\
26&\ding{51}  & \ding{55} & \ding{51} & -- & \ding{51} &3 &62&0 &0 &62 &--\\
27&\ding{51} & \ding{51}& \ding{51} & -- & \ding{55}&3&38&11 &0 &49 &--\\
28&\ding{55} & \ding{51}& \ding{51} & -- & --&3&88& 0 &0 &88 &--\\
29&\ding{55} &  \ding{51} & -- & -- &  \ding{55} &2 &27&0 &0 &27 &--\\
30&\ding{51} & \ding{55}& \ding{55} & -- &\ding{51} &2 &25&0 &0 &25 &--\\
31&\ding{55} & \ding{55} & -- & -- & \ding{51}&1 &30& 0&0 &30 &--\\
32&\ding{55}&   \ding{55}& \ding{55} & \ding{55} & \ding{55} &1 &12&0  &0 &12 &--\\
33&\ding{51} & \ding{51} & \ding{51} & -- & \ding{51} & 1 & 0 & 0 & 21 & 21 &0\\
34&\ding{55} & \ding{55} & \ding{51} & -- & \ding{55} & 1 & 0 & 0 & 15 & 15 &0\\
35&\ding{51} & \ding{55} & \ding{55} & \ding{51} & \ding{55} & 1 & 0 & 0 & 53 & 53 &0\\
36&\ding{51}&   \ding{55}  & \ding{55}& \ding{55}  & \ding{55} &1 &0&0  &31 &31 &0\\
\hline
& & & & &  &10,180 &240,013 &35,182 & 64,589 &339,784 & \textbf{weighted mAP: 72.3}\\
\end{tabular}
\begin{tablenotes}
\item[\ding{51}] rider in corresponding position wears a helmet
\item[\ding{55}] rider in corresponding position does not wear a helmet
\item[--] there is no rider in corresponding position
\end{tablenotes}
\end{threeparttable}
}
\label{tb:instancestat_and_accuracy}
\end{table}

\newpage
\section{Comparison to human observation in real world application} \label{human}
Since the video data that forms the basis for the training of the machine learning algorithm in this paper has been analyzed in the past to assess motorcycle helmet use, there is a unique opportunity to compare hand-counted helmet use numbers in the video data with the calculated helmet use numbers generated by the algorithm developed in this paper. Siebert et al. \citep{siebert2019patterns} hand-counted the motorcyclists with and without helmets in the source video data for the first 15 minutes of every hour that a video was recorded. Hence, "hourly" helmet use percentages for every individual observation site in the data set are available. To assess the feasibility of our machine learning approach for real-world observation studies, we compare hourly hand-counted helmet use rates from the Siebert et al. study with hourly computer-counted rates estimated through the application of our algorithm. 

\subsection{Method} \label{Method_human}
It is important to understand the fundamental difference of the hand-counting method used by Siebert et al. \cite{siebert2019patterns} and the frame-based algorithmic approach presented in this paper. In the initial observation of the video data by Siebert et al., a human observer screened 15 minute video sections and registered the number of helmet- and non-helmet-wearing motorcycle riders for individual motorcycles. I.e. helmet use on a motorcycle was only registered once, even though an individual motorcycle was present in multiple frames, driving through the field of view of the camera. The occlusion of an individual motorcycle in some of these frames, e.g. when a motorcycles passed a bus that was located between the motorcycle and the camera, does not pose a problem for the detection of helmet use on that individual motorcycle, as the human observer has the possibility to jump back and forth in the video and register helmet use in a frame with a clear unoccluded view of the motorcycle. Furthermore, a human observer naturally tracks a motorcycle and can easily identify a frame where the riders of the motorcycle and their helmet use is most clearly visible, e.g. when the motorcycle is closest to the camera. The human observer can then use this frame to arrive at a conclusion on the number of riders and their helmet use.

In contrast, the computer vision approach developed in this paper will register motorcycle riders' helmet use in each frame where a motorcycle is detected. This can introduce some error-variance in helmet use detection. The speed of a motorcycle will influence how many times helmet use for an individual motorcycle is registered, as slower motorcycle riders will appear in more frames than faster ones. Furthermore, occlusion influences how many times a motorcycle will be registered, which can influence the overall helmet use average calculated. Also, helmet use will be registered for motorcycles that are in a sub-optimal angle to the camera, e.g. on motorcycles that drive directly towards the camera, drivers can occlude passengers behind them. However, not all of these differences have a direct impact on helmet use calculated through the algorithm. We assume that occlusion does not introduce a directed bias to detected helmet use rates, as riders with and without helmets have the same chance to be occluded by other traffic. The same can be assumed for differences in motorcycle speed within the observed cities, as riders with helmets won't be faster or slower than those without helmets. We therefore assume that a frame based helmet use registration will lead to comparable results to helmet use registered by a human observer.

Since the algorithm has been trained on specific observation sites, it can be considered to be \textit{observation site trained}. I.e. when the algorithm is applied to the observation site \texttt{Bago\_rural}, there is data on this specific observation site in the training set (Table \ref{tb:dataset}). In an application of the deep learning based approach, this might not be the case, as the algorithm will not have been trained on new observation sites. Hence, in the following, we also compare algorithmic accuracy for an \textit{observation site untrained} algorithm. For this, we exclude all training data from the observation site that we analyze, before training the algorithm, simulating the application of the algorithm to a new observation site. In the following, \textit{trained algorithm} refers to the algorithm with training on an observation site to be analyzed, while \textit{untrained algorithm} refers to an algorithm that was not trained on an observation site to be analyzed.

\subsection{Results} \label{Results_human}
\begin{figure*}[hbt!]
    \centering
    \begin{subfigure}[b]{0.33\textwidth}
        \includegraphics[width=0.95\textwidth]{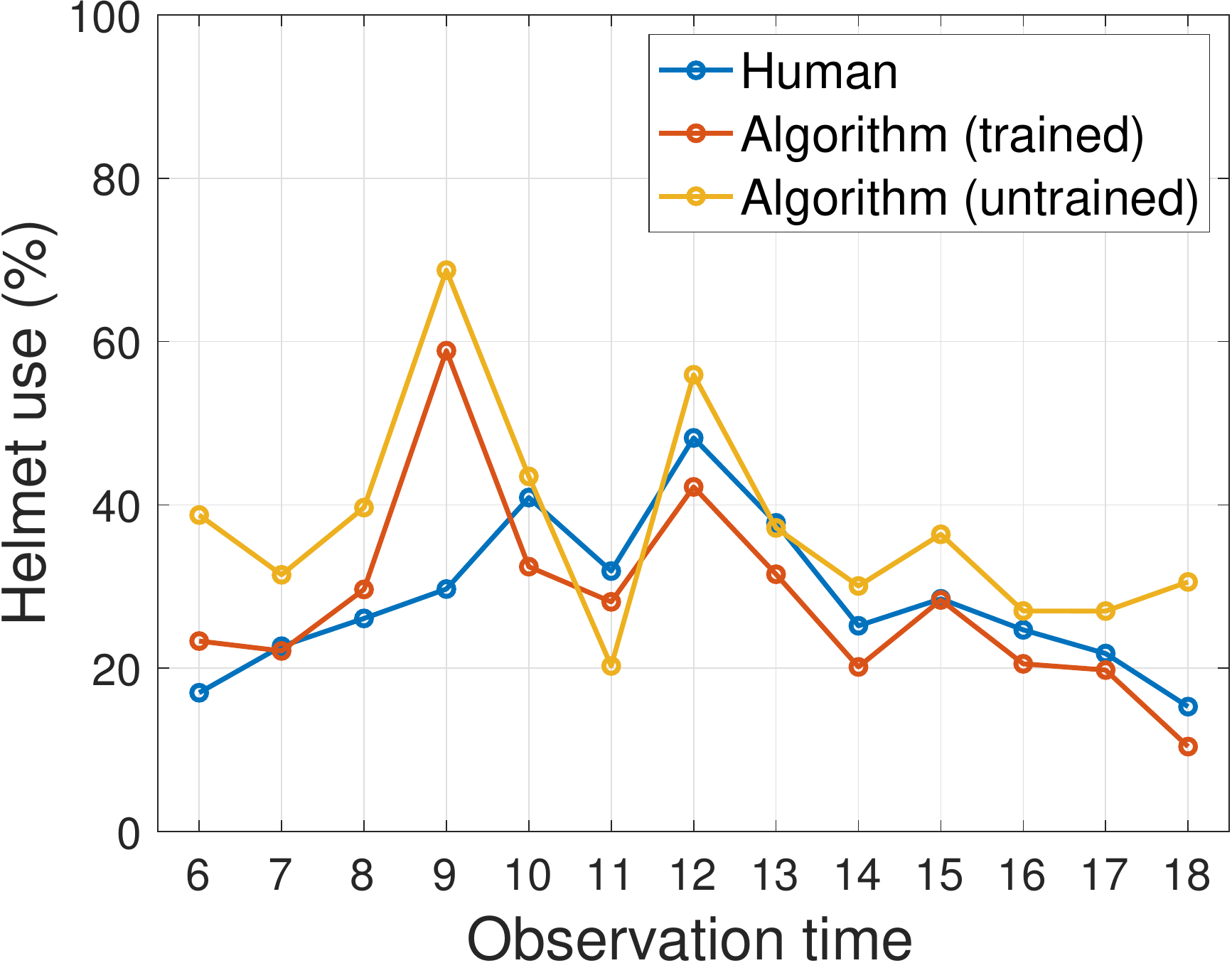}
        \caption{\texttt{Bago\_rural}}
    \end{subfigure}
    \begin{subfigure}[b]{0.33\textwidth}
        \includegraphics[width=0.95\textwidth]{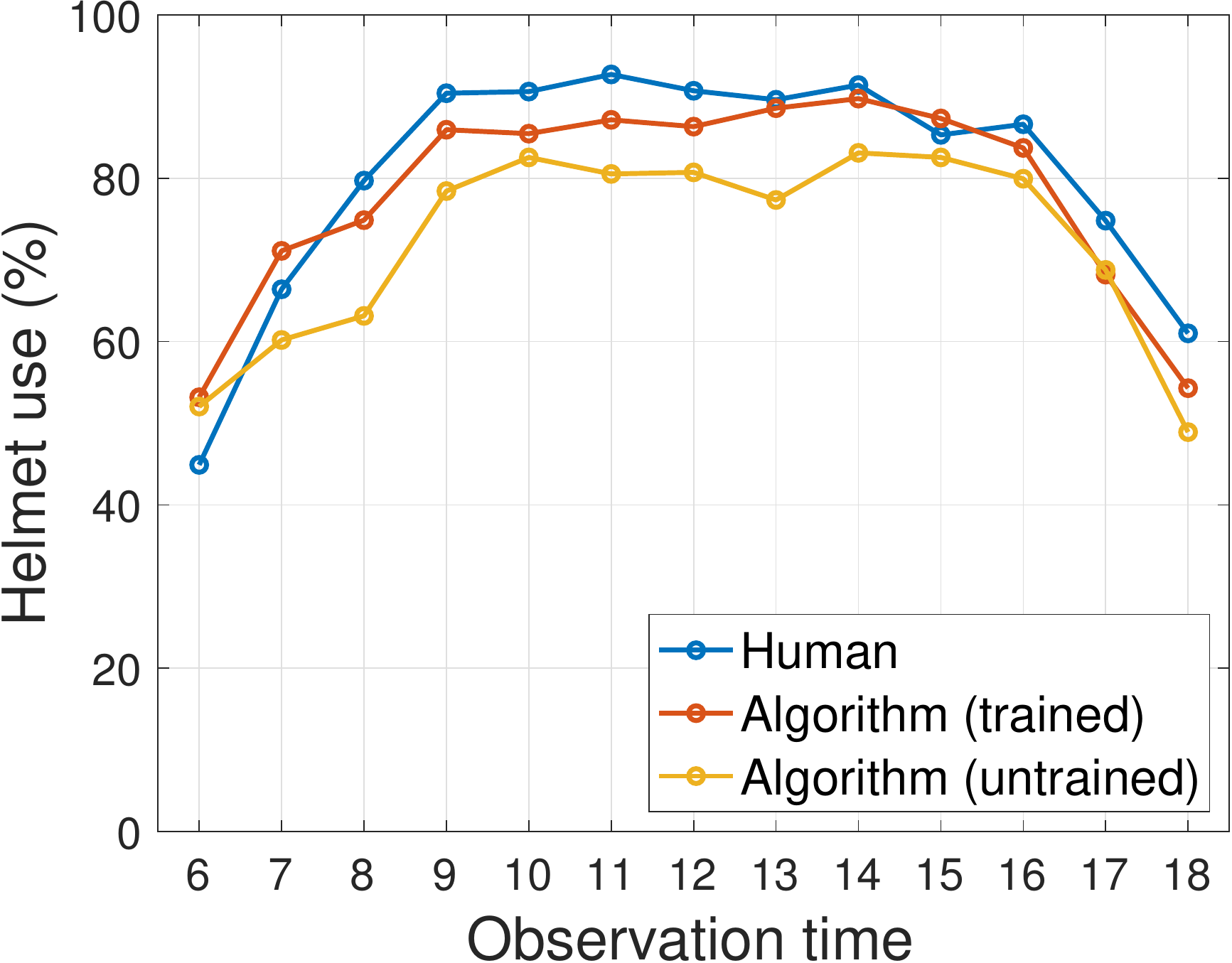}
        \caption{\texttt{Mandalay\_1}}
    \end{subfigure}    
    \begin{subfigure}[b]{0.33\textwidth}
        \includegraphics[width=0.95\textwidth]{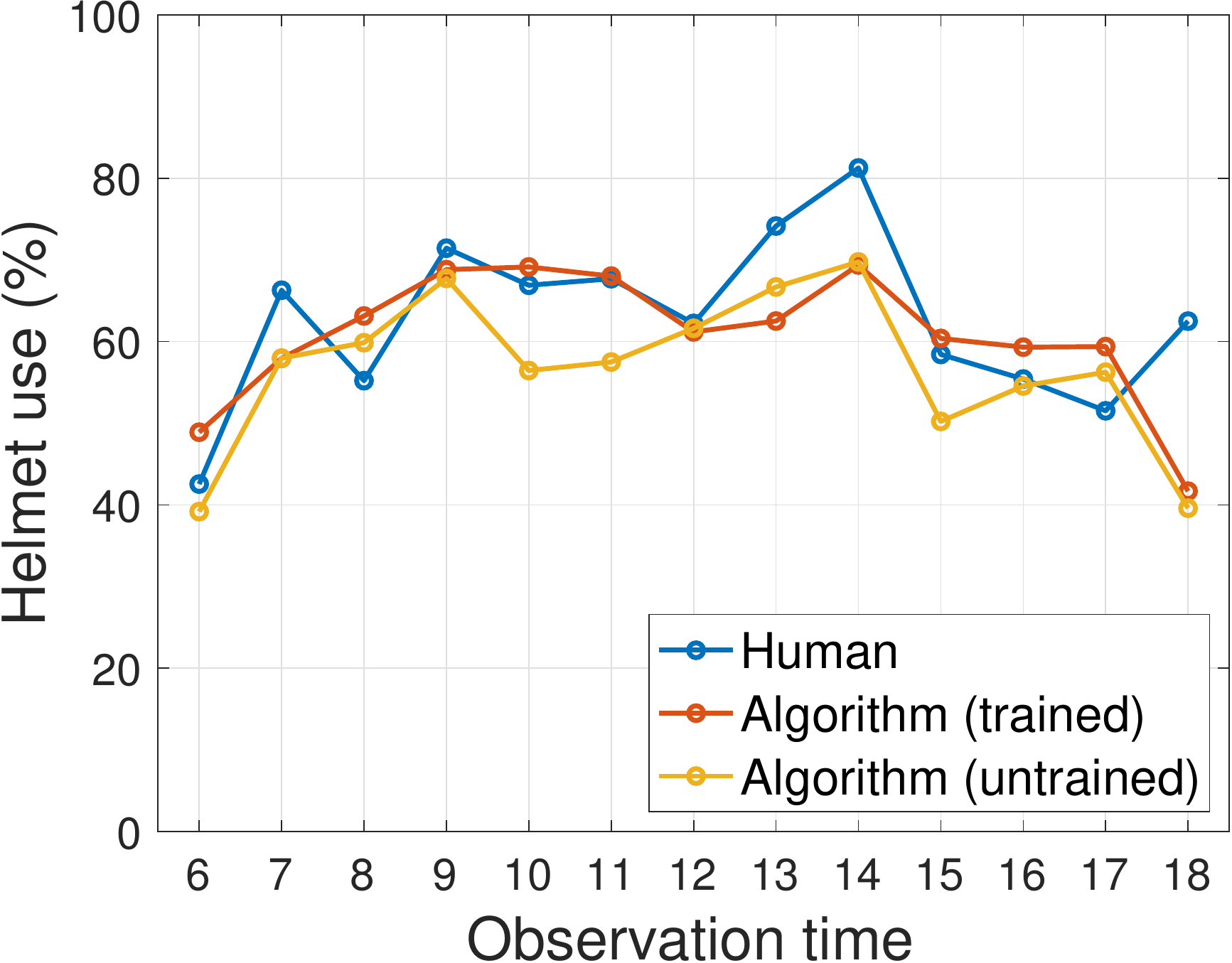}
        \caption{\texttt{Naypitaw\_2}}
    \end{subfigure}
    \begin{subfigure}[b]{0.33\textwidth}
        \includegraphics[width=0.95\textwidth]{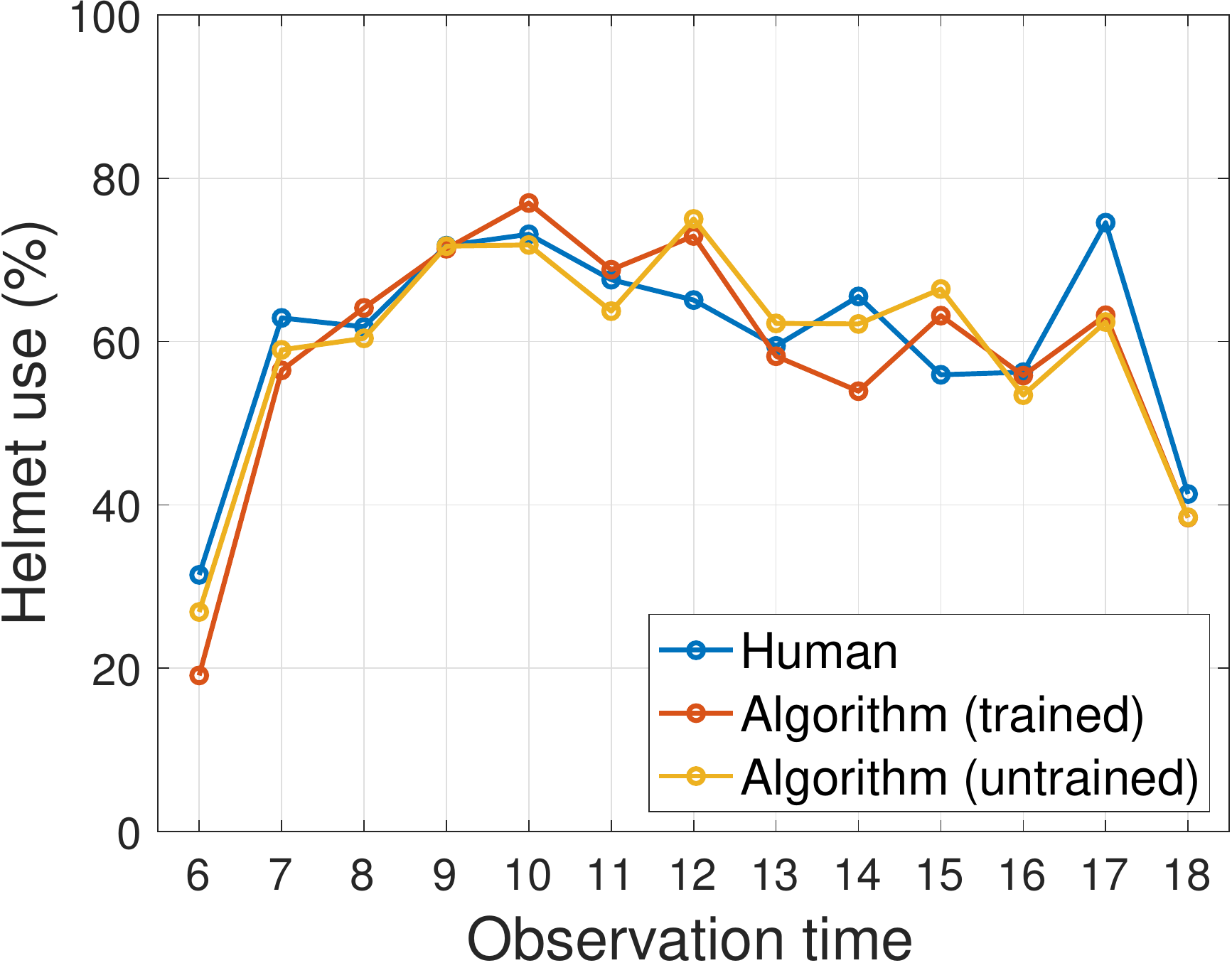}
        \caption{\texttt{NyaungU\_rural}}
    \end{subfigure}
    \begin{subfigure}[b]{0.33\textwidth}
        \includegraphics[width=0.95\textwidth]{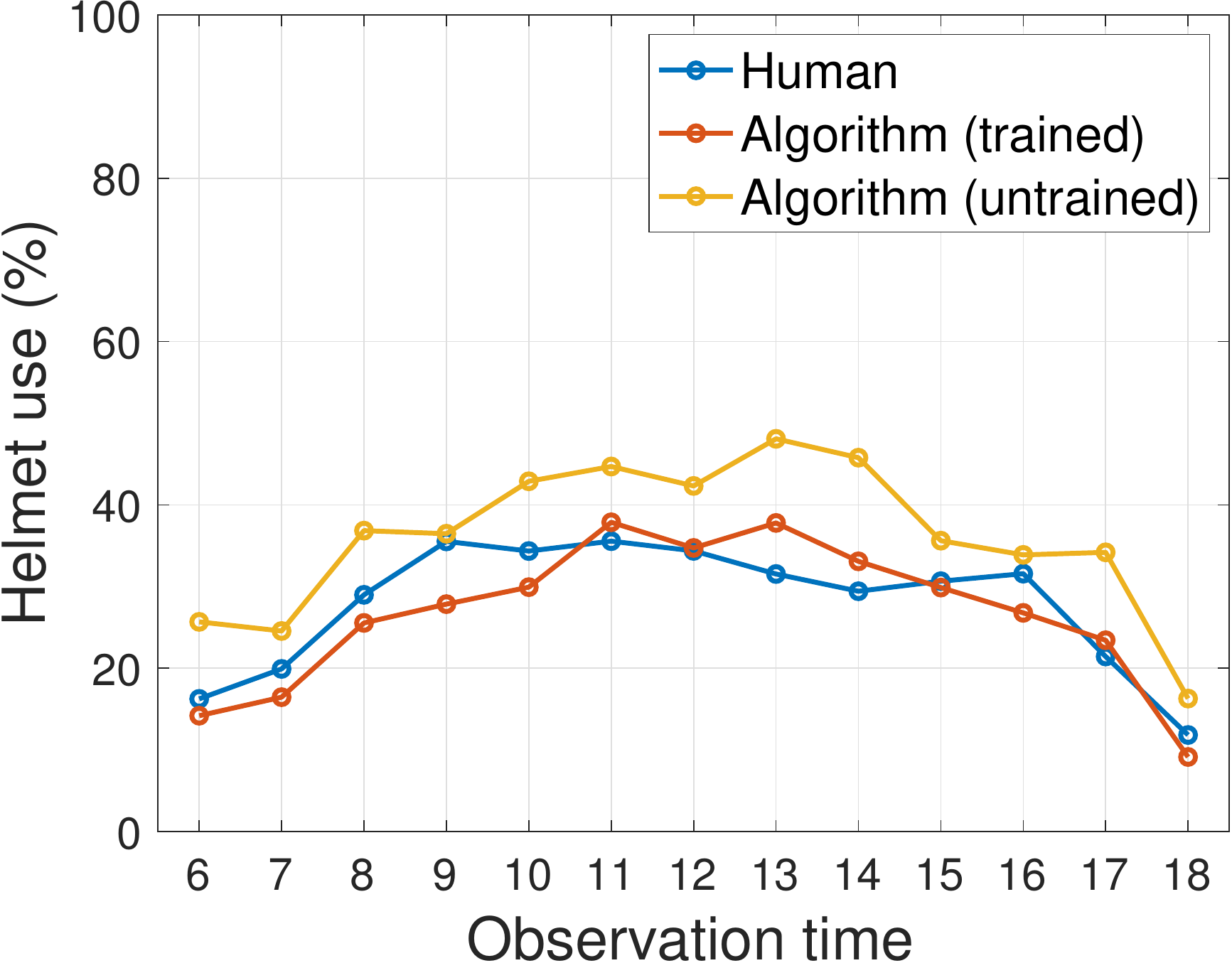}
        \caption{\texttt{Pakokku\_urban}}
    \end{subfigure}
    \begin{subfigure}[b]{0.33\textwidth}
        \includegraphics[width=0.95\textwidth]{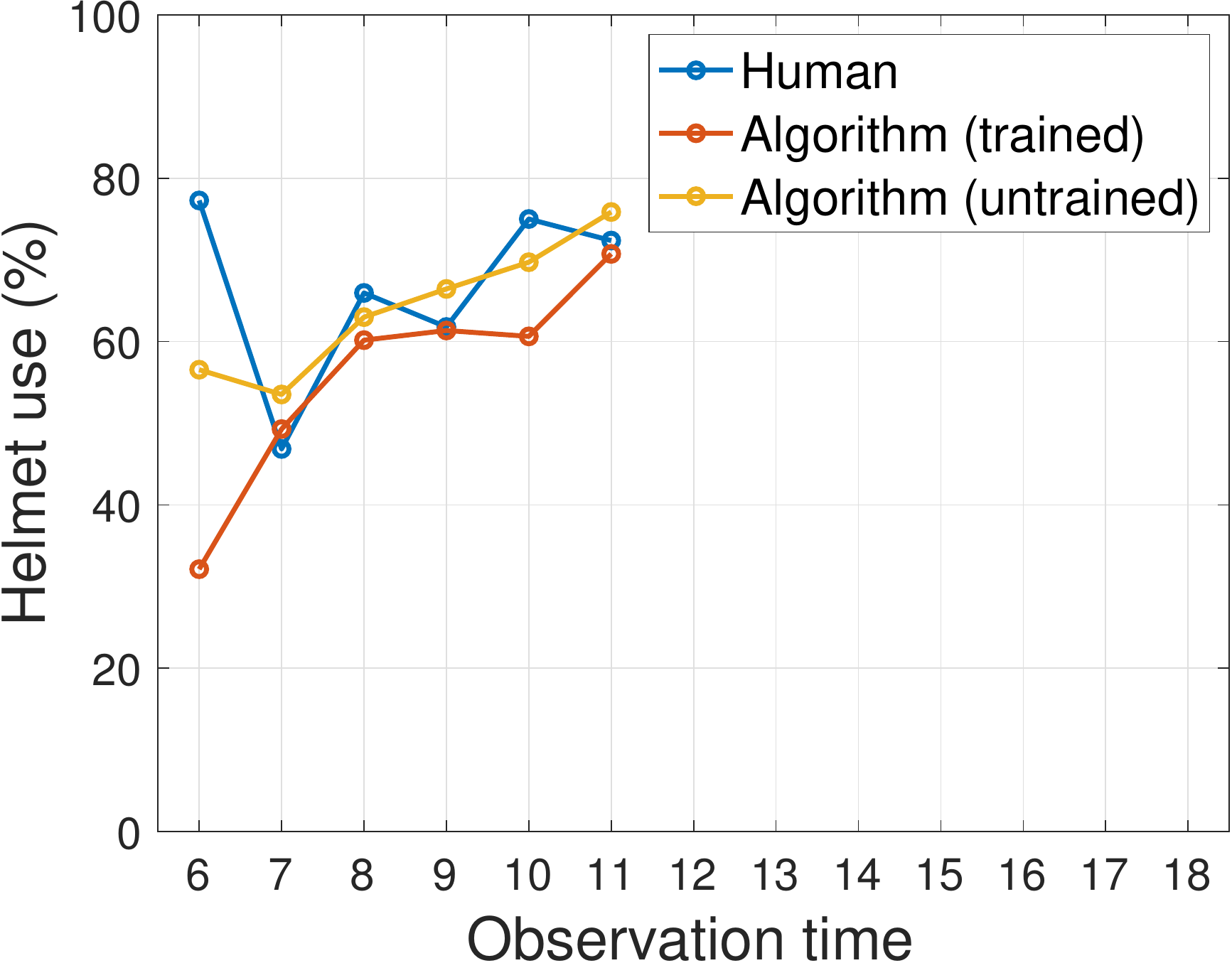}
        \caption{\texttt{Pathein\_rural}}
    \end{subfigure}
    \begin{subfigure}[b]{0.33\textwidth}
        \includegraphics[width=0.95\textwidth]{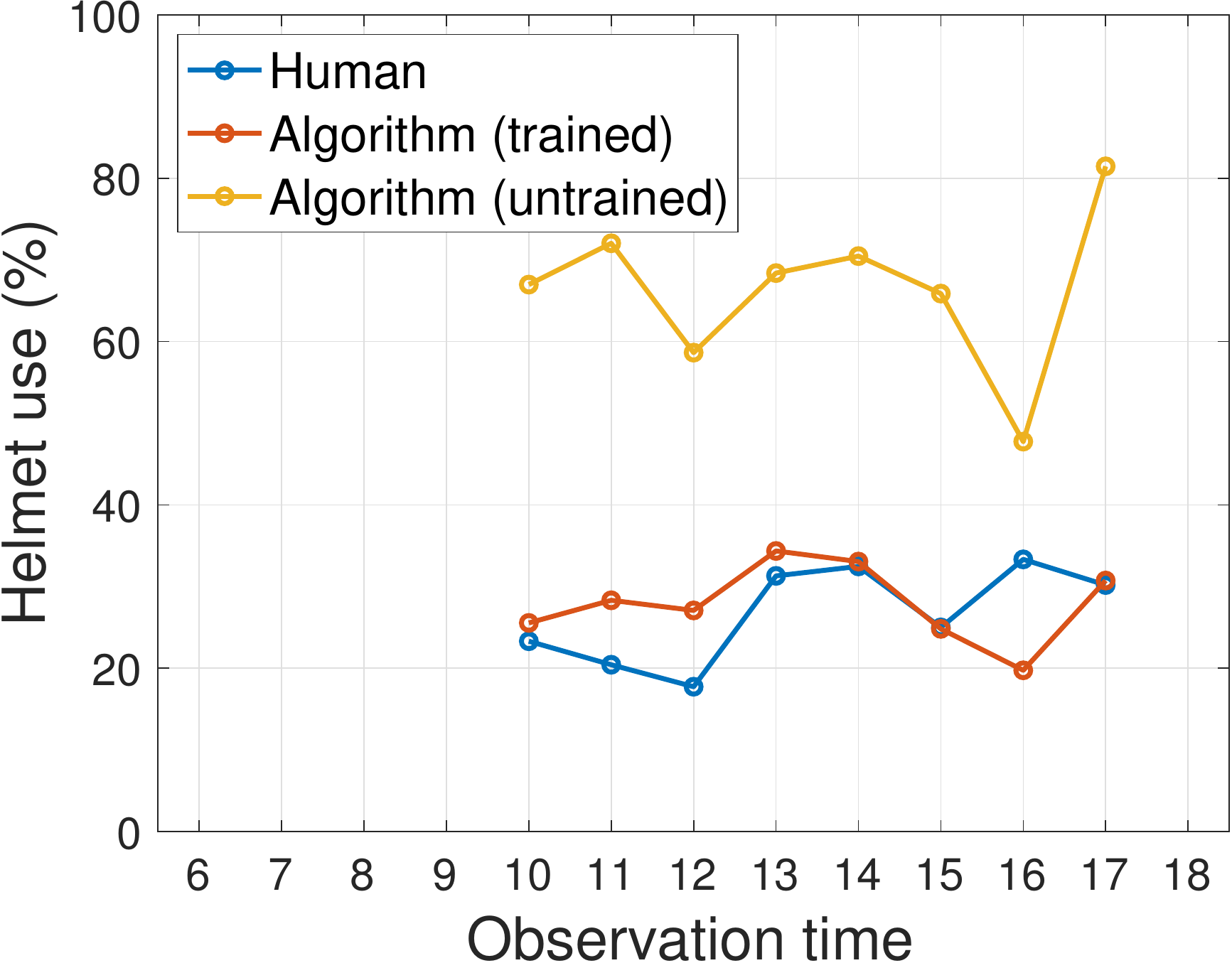}
        \caption{\texttt{Yangon\_II}}
    \end{subfigure}
    \caption{Hourly helmet use averages for one day of each observation site, registered by human observers and the \textit{trained} and \textit{untrained} algorithm (incomplete data for \texttt{Pathein\_rural} and \texttt{Yangon\_II} due to technical problems during the video data collection).}
    \label{fig:human_vs_machine_hourly}
\end{figure*}

Data on hourly helmet use rates for one randomly chosen day of video data from each observation site is presented in Fig.~\ref{fig:human_vs_machine_hourly}. Helmet use was either registered by a human observer \citep{siebert2019patterns}, registered through the \textit{trained} algorithm, or the \textit{untrained} algorithm. It can be observed that hourly helmet use percentages are relatively similar when comparing human and computer registered rates of the trained algorithm. The trained algorithm registers accurate helmet use rates, even when large hourly fluctuations in helmet use are present, e.g. at the \texttt{Mandalay\_1} observation site (Figure \ref{fig:human_vs_machine_hourly}(b)). However, some of the observed 15 minute videos show a large discrepancy between helmet use rates registered by a human and the trained algorithm. While it is not possible to conduct a detailed error analysis (as we did in Section \ref{Results_ML}) it is possible to evaluate the video data for broad factors that could increase the discrepancy between human registered data and the data registered by the trained algorithm.

\begin{figure*}[htb!]
    \centering
    \begin{subfigure}[b]{0.45\textwidth}\centering
        \includegraphics[width=1.0\textwidth]{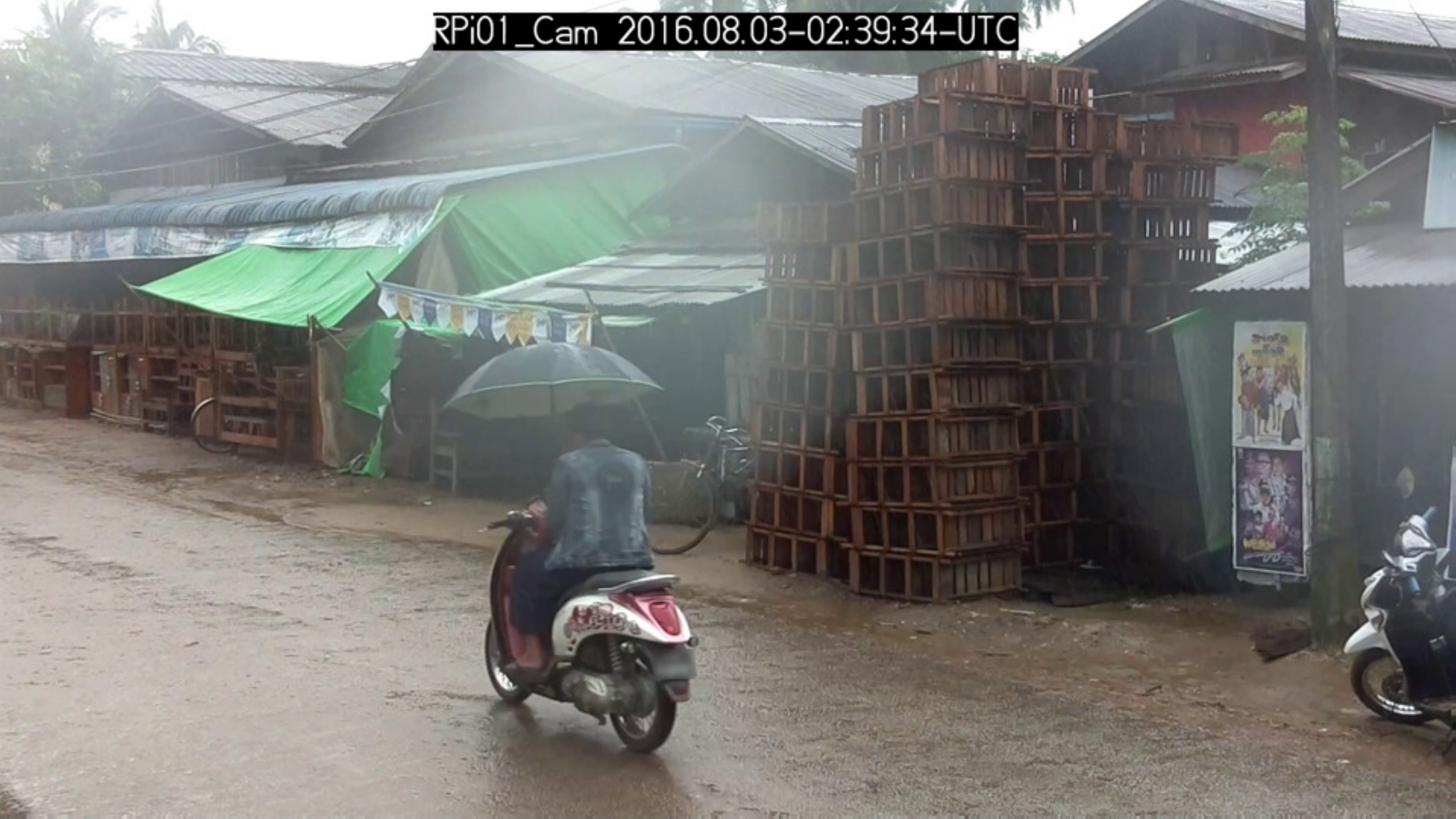}
    \end{subfigure}
    \hspace{1pt}
    \begin{subfigure}[b]{0.45\textwidth}\centering
        \includegraphics[width=1.0\textwidth]{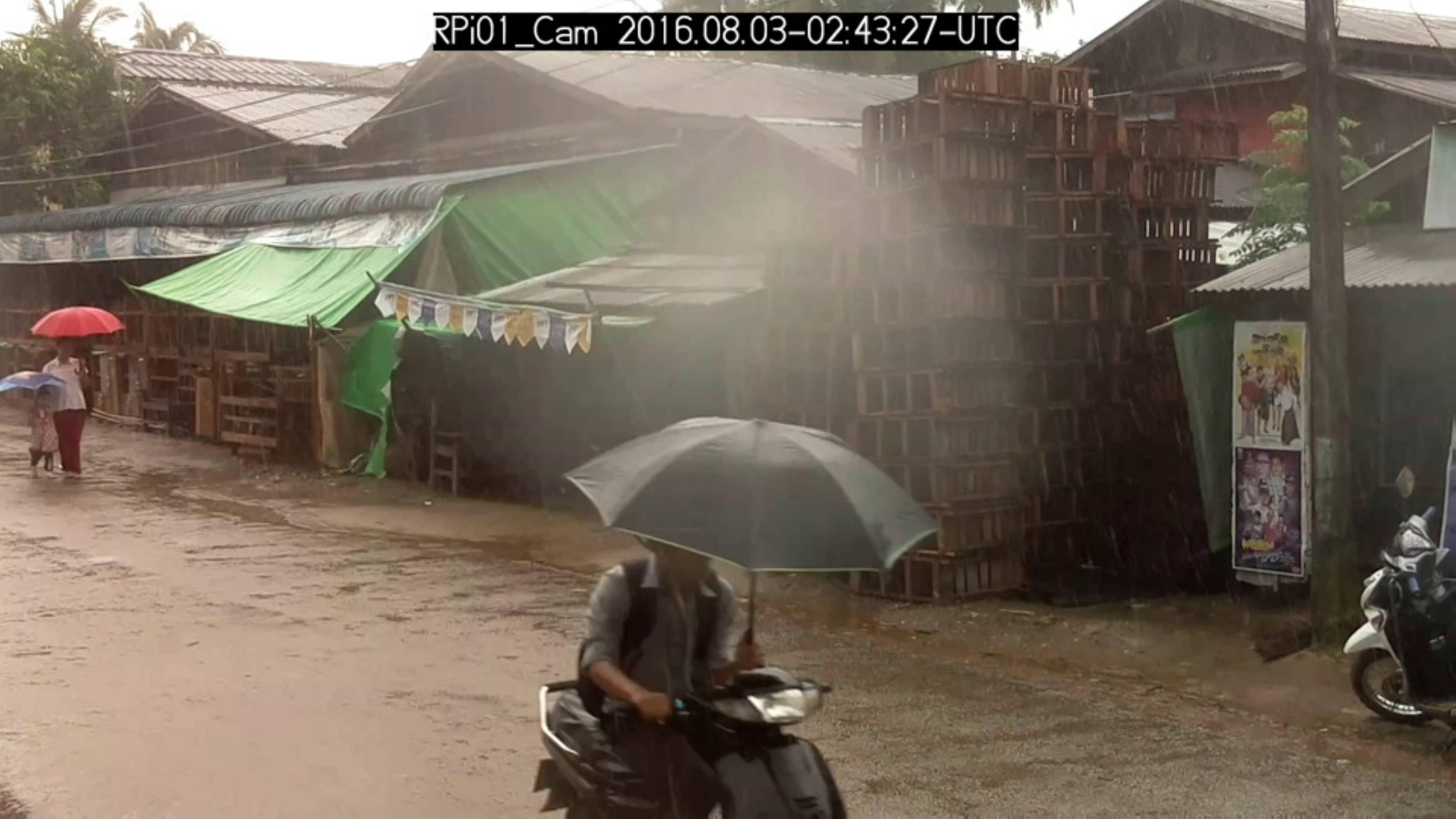}
    \end{subfigure}
    
    \caption{Video frames from the Bago observation site at 9 am. Heavy rain, fogging on the camera lens, and umbrella use is visible.}\label{fig:bago_rain}
\end{figure*}

As an example, helmet use rates at the \texttt{Bago\_rural} observation site at 9 am have a much higher helmet use rate registered by the trained algorithm than by the human observer (Figure \ref{fig:human_vs_machine_hourly}(a)). A look at the video data from this time frame reveals heavy rain at the observation site (Fig.~\ref{fig:bago_rain}). Apart from an increased blurriness of frames due to a decrease in lighting and visible fogging on the inside of the camera case, motorcycle riders can be observed to use umbrellas to protect themselves against the rain. It can be assumed that motorcycle riders without helmets are more likely to use an umbrella, as they are not protected from the rain by a helmet. This could explain the higher helmet use registered by the trained algorithm at this observation site and time, as non-helmeted riders are less likely to be detected due to umbrellas.

\begin{figure*}[!htb]
    \centering
    \begin{subfigure}[b]{0.45\textwidth}\centering
        \includegraphics[width=1.0\textwidth]{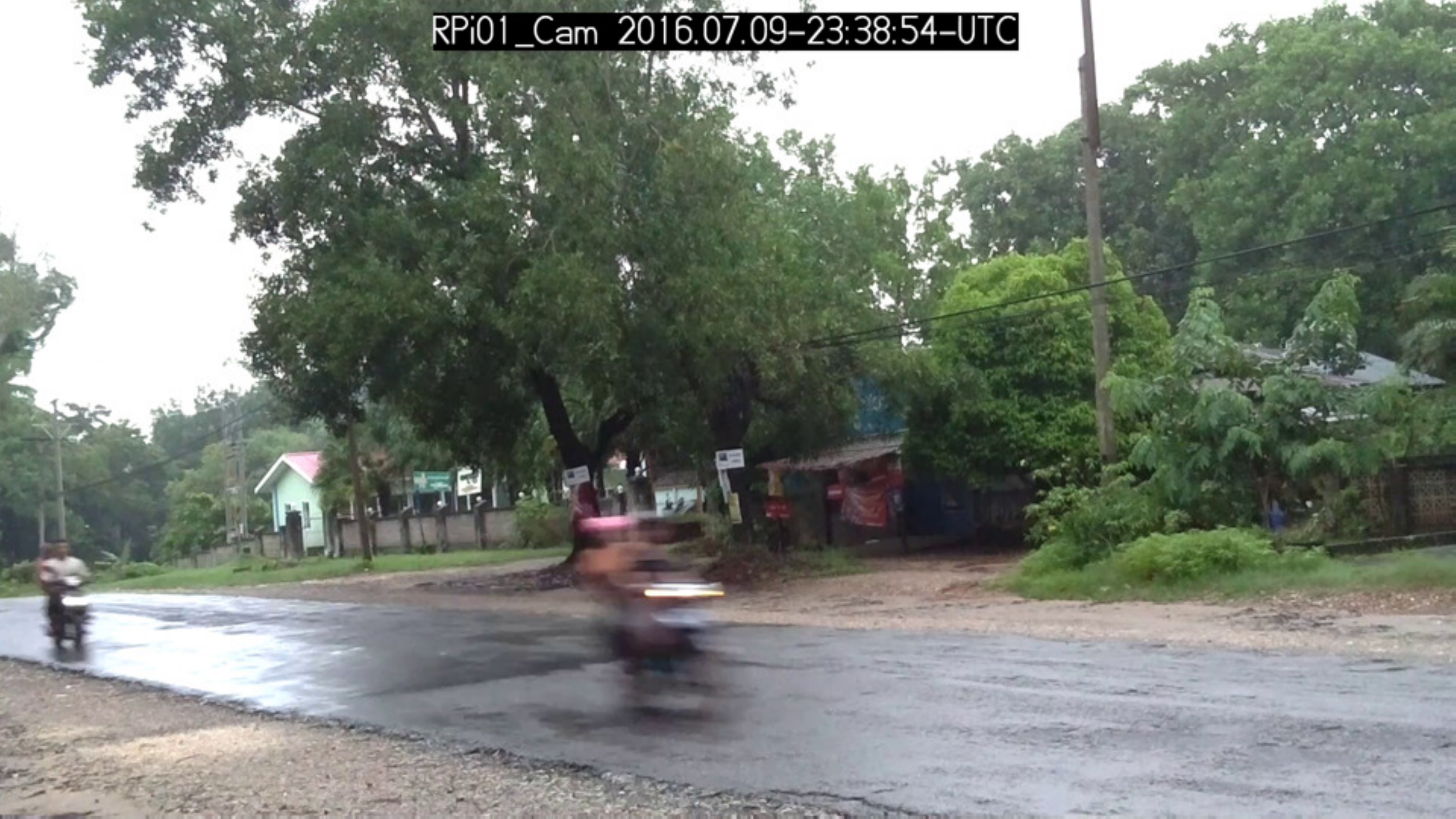}
    \end{subfigure}
    
    \caption{A video frames from the Pathein observation site at 6 am. Low lighting due to heavy clouds results in blurry motorcycles.}\label{fig:pathein_clouds}
\end{figure*}

Another instance of a large discrepancy between human and computer registered helmet rates through the trained algorithm can be observed for 6 am at the \texttt{Pathein\_rural} observation site (Figure \ref{fig:human_vs_machine_hourly}(f)). A look at the video data reveals bad lighting conditions due to a combination of clouded weather and the early observation time. This results in unclear motorcycles, which are blurred due to their driving speed in combination with the bad lighting conditions (Fig.~\ref{fig:pathein_clouds}).

\begin{figure*}[!htb]
    \centering
    \begin{subfigure}[b]{0.6\textwidth}\centering
        \includegraphics[width=1.0\textwidth]{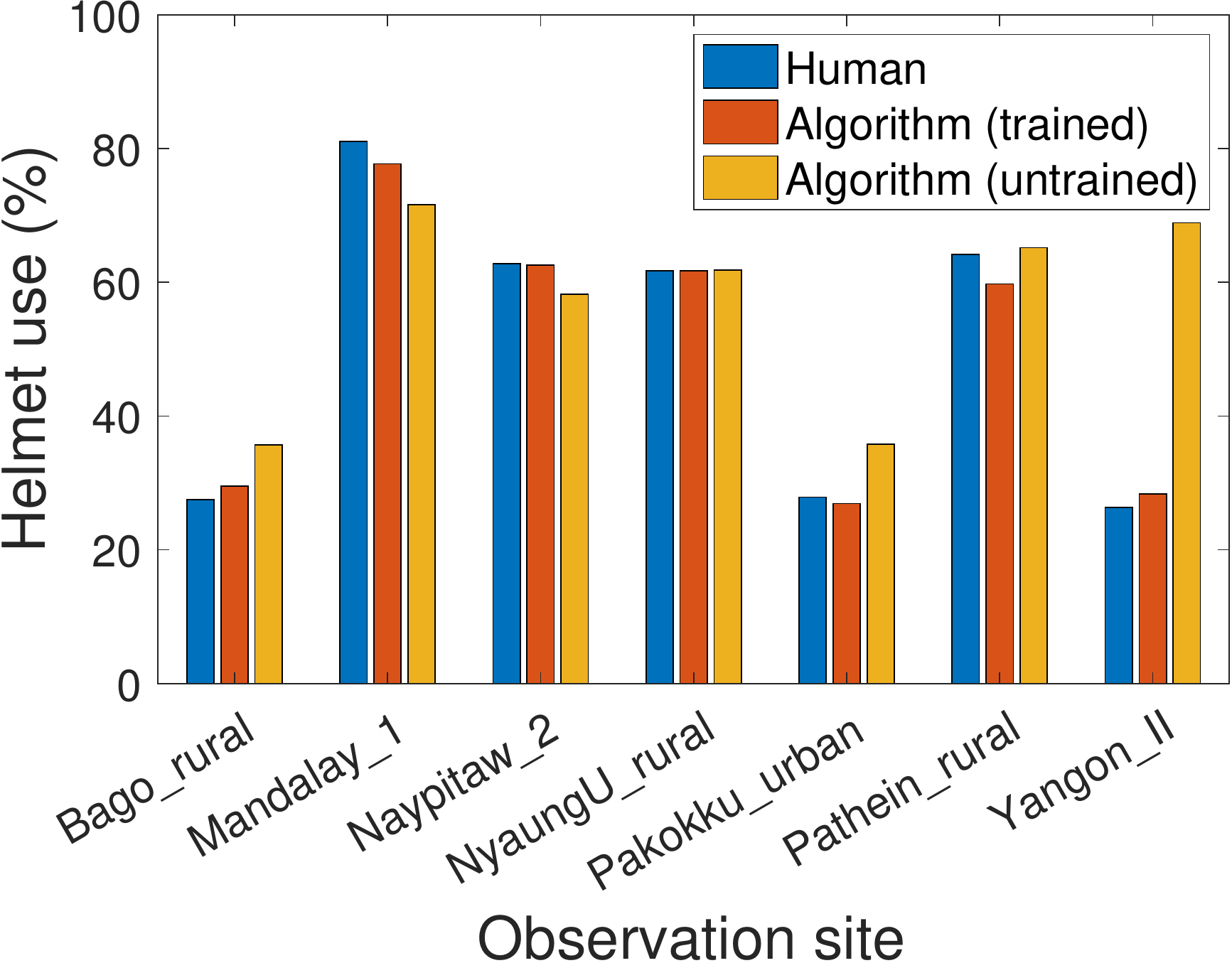}
    \end{subfigure}
    \caption{Average helmet use percentage registered by a human observer and the trained and untrained algorithm.}\label{fig:human_vs_machine_site}
\end{figure*}

Despite singular discrepancies between hourly helmet use rates coded by a human and the trained algorithm, the overall accuracy of average helmet use rates calculated by the trained algorithm per observation site is high. A comparison of average helmet use per observation site, registered by a human observer and the trained algorithm is presented in (Figure \ref{fig:human_vs_machine_site}). For three observation sites (Naypitaw, Nyaung-U, and Pakokku), trained algorithm registered helmet use rates deviate by a maximum of 1\% from human registered rates. For the other four observation sites (Bago, Mandalay, Pathein, and Yangon), trained algorithm registered rates are still reasonably accurate, varying between -4.4\% and +2.07\%.

\begin{figure*}[!htb]
    \centering
    \begin{subfigure}[b]{1.0\textwidth}\centering
        \includegraphics[width=1.0\textwidth]{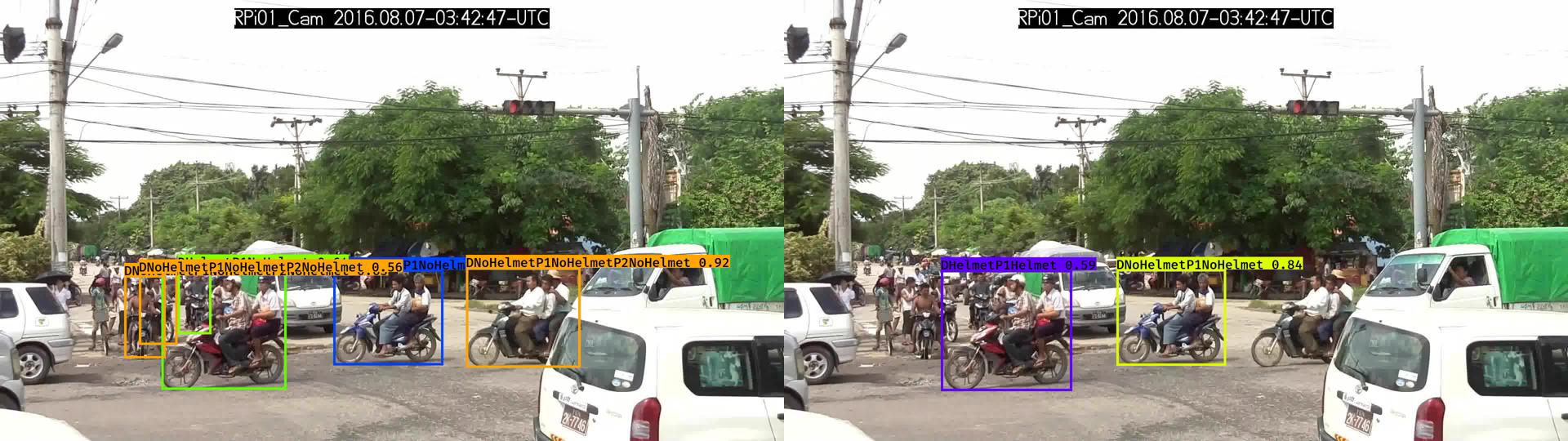}
    \end{subfigure}
    
    \caption{Comparison of the trained (left) and untrained algorithm (right) at the \texttt{Yangon\_II} observation site.}\label{fig:Yangon_II}
\end{figure*}

For the untrained algorithm, it can be observed that the registered hourly helmet use data is less accurate than that of the trained algorithm, while it is still relatively close to the human registered data for most observation sites (\mbox{Figure \ref{fig:human_vs_machine_hourly})}. The effects of decreased visibility at the \texttt{Bago\_rural} and \texttt{Pathein\_rural} sites are also present. However, at the \texttt{Yangon\_II} observation site, the registered helmet use of the untrained algorithm is notably higher than helmet use registered by the trained algorithm and the human observer, registering more than double the helmet use present at the observation site. A comparison of the frame-level helmet use detection at the \texttt{Yangon\_II} site between the trained and untrained algorithm revealed a large number of missed detections of the untrained algorithm (Figure \ref{fig:Yangon_II}). Excluding \texttt{Yangon\_II}, the helmet use rates registered through the untrained algorithm vary between -8.13\% and +9.43\% from human registered helmet use (Figure \ref{fig:human_vs_machine_site}).

\section{Discussion}
In this paper, we set out to develop a deep learning based approach to detect motorcycle helmet use. Using a large number of video frames we trained an algorithm to detect active motorcycles, the number and position of riders, as well as their helmet use. The use of an annotated test data set allowed us to evaluate the accuracy of our algorithm in detail (Section \ref{Results_ML}, Table \ref{tb:instancestat_and_accuracy}). The algorithm had high accuracy for the general detection of motorcycles. Further, it was capable of accurately identifying the number of riders and their position on the motorcycle. The algorithm was less accurate however, for motorcycles with a large number of riders or for motorcycles with an uncommon rider composition (Table \ref{tb:instancestat_and_accuracy}). Based on these results, the present version of the algorithm can be expected to generate highly accurate results in countries, where only two riders are allowed on a motorcycle and where riders adherence to this law is high. Our implementation of the algorithm can run on consumer hardware with a speed of 14 frames per second, which is higher than the frame rate of the recorded video data. Hence, the algorithm can be implemented to produce real-time helmet use data at any given observation site.

Our comparison of algorithm accuracy with helmet use registered by a human observer (Section \ref{human}) revealed an overall high average accuracy, if the algorithm had been trained on the specific observation sites (Figure \ref{fig:human_vs_machine_site}). If there was no prior training on the specific observation site, the (untrained) algorithm had an overall lower accuracy in helmet use detection. There was a large deviation of registered helmet use at the \texttt{Yangon\_II} observation site, where a large number of missed detections resulted in a highly inaccurate detection performance. The lack of training data with a camera angle similar to the \texttt{Yangon\_II} observation site is the most likely cause for this low detection accuracy. Potential ways to counteract this performance decrement are discussed in the Section \ref{Conclusion}.

A comparison of hourly helmet use rates revealed a small number of discrepancies between human and algorithm registered rates (Fig. \ref{fig:human_vs_machine_hourly}). Further analysis revealed a temporary decrease in the video source material quality as the reason for these discrepancies (Fig. \ref{fig:bago_rain} \& \ref{fig:pathein_clouds}). This decrease in detection accuracy has to be seen in light of the training of the algorithm, in which periods with motion blur due to bad lighting or bad weather were excluded. Hence, decrements in detection accuracy are not necessarily the result of differences in observation sites themselves.

\section{Conclusion and future work} \label{Conclusion}


The lack of representative motorcycle helmet use data is a serious global concern for governments and road safety actors. Automated helmet use detection for motorcycle riders is a promising approach to efficiently collect large, up-to-date data on this crucial measure. When trained, the algorithm presented in this paper can be directly implemented in existing road traffic surveillance infrastructure to produce real-time helmet use data. Our evaluation of the algorithm confirms a high accuracy of helmet use data, that only deviates by a small margin from comparable data collected by human observers. Observation site specific training of the algorithm does not involve extensive data annotation, as already the annotation of 270 s of video data is enough to produce accurate results for e.g. the \texttt{Yangon\_II} observation site. While the sole collection of data does not increase road safety by itself \citep{hyder2019measurement}, it is a prerequisite for targeted enforcement and education campaigns, which can lower the rate of injuries and fatalities \citep{world2006helmets}.

For future work we propose three ways in which the software-side performance of machine learning based motorcycle helmet use detection can be improved. First, there is a need to collect more data in under-represented classes (Table \ref{tb:instancestat_and_accuracy}) to increase rider, position, and helmet detection accuracy for motorcycles with more than two riders. Second, diverse video data should be collected in regards to the camera angle. This would prevent detection inaccuracies caused by missed detections in camera setups with unusual camera angles. Third, it appears promising to add a simple tracking method for motorcycles to the existing approach. Tracking would allow the identification of individual motorcycles within a number of subsequent frames. Using a frame based quality assessment of an individual motorcycle's frames together with tracking, would allow the algorithm to choose the most suitable frame for helmet use and rider position detection, which will improve overall detection accuracy. Tracking would further allow the algorithm to register the number of individual motorcycles passing an observation site, providing valuable information on traffic flows and density.

On the hardware-side, future applications of the algorithm can greatly benefit from an improved camera system, that is less influenced by low light conditions (Fig. \ref{fig:pathein_clouds}) and less susceptible to fogging and blur due to rain on the camera lens (Fig. \ref{fig:bago_rain}). An increase of the resolution of the video data could allow the detection of additional measures, such as helmet type or chin-strap usage. Apart from a generally increased performance through software and hardware changes, future applications of the developed method could incorporate a more comprehensive set of variables. Within the deep learning approach, the detection of e.g. age categories, chin-strap use, helmet type, or mobile phone use would be possible.

There are a number of limitations to this study. Algorithmic accuracy was only analyzed for road environments within Myanmar, limiting the type of motorcycles and helmets present in the training set. Future studies will need to assess whether the algorithm can maintain the overall high accuracy in road environments in other countries. A similar limitation can be seen in the position of the observation camera. While the algorithm is able to detect motorcycles from a broad range of angles due to diverse training data, there was no observation site where the observation camera was installed in an overhead position, filming traffic from above. Since traffic surveillance infrastructure is often installed at this position, future studies will need to assess whether the algorithm would produce accurate results from an overhead angle. This is especially important in light of the results of the \texttt{Yangon\_II} observation site, where an unusual camera angle lead to a large number of missed detections. Furthermore, a more structured variation of camera to lane angle would help to better understand optimal positioning of observation equipment for maximum detection accuracy.
While it was included in the data annotation process, the algorithmic accuracy in detecting the position of riders was not compared to human registered data in this study. In light of large differences of motorcycle helmet use for different rider positions \citep{siebert2019patterns}, future studies will need to incorporate deeper analysis of position detection accuracy. For the comparison of human- and machine-registered helmet use rates, it appears promising to enable a detailed error analysis (false positive/ false negative) through the use of an adapted data structure of human helmet use registration.   


In conclusion, we are confident that automated helmet use detection can solve the challenges of costly and time-consuming data collection by human observers. We believe that the algorithm can facilitate broad helmet use data collection and encourage its active use by actors in the road safety field.

\section*{Acknowledgement}
This research was supported by the \textit{Deutsche Forschungsgemeinschaft} (DFG, German Research Foundation) – Project-ID 251654672 – TRR 161 (Project A05).

\newpage
\bibliography{refs}

\begin{thebibliography}{10}
\expandafter\ifx\csname url\endcsname\relax
  \def\url#1{\texttt{#1}}\fi
\expandafter\ifx\csname urlprefix\endcsname\relax\def\urlprefix{URL }\fi
\expandafter\ifx\csname href\endcsname\relax
  \def\href#1#2{#2} \def\path#1{#1}\fi

\bibitem{liu2004helmets}
B.~Liu, R.~Ivers, R.~Norton, S.~Blows, S.~K. Lo, Helmets for preventing injury
  in motorcycle riders, Cochrane database of systematic reviews 4 (2004) 1--42.

\bibitem{bachani2013trends}
A.~M. Bachani, C.~Branching, C.~Ear, D.~R. Roehler, E.~M. Parker, S.~Tum, M.~F.
  Ballesteros, A.~A. Hyder, Trends in prevalence, knowledge, attitudes, and
  practices of helmet use in cambodia: results from a two year study, Injury 44
  (2013) S31--S37.

\bibitem{bachani2017helmet}
A.~Bachani, Y.~Hung, S.~Mogere, D.~Akunga, J.~Nyamari, A.~A. Hyder, Helmet
  wearing in kenya: prevalence, knowledge, attitude, practice and implications,
  Public health 144 (2017) S23--S31.

\bibitem{siebert2019patterns}
F.~W. Siebert, D.~Albers, U.~Aung~Naing, P.~Perego, S.~Chamaiparn, Patterns of
  motorcycle helmet use – a naturalistic observation study in myanmar,
  Accident Analysis \& Prevention 124 (2019) 146--150.

\bibitem{world2015global}
{World Health Organization}, Global status report on road safety 2015, World
  Health Organization, 2015.

\bibitem{fong2015smallsample}
M.~C. Fong, J.~R. Measelle, J.~L. Dwyer, Y.~K. Taylor, A.~Mobasser, T.~M.
  Strong, S.~Werner, S.~Ouansavanh, A.~Mounmingkham, M.~Kasuavang, et~al.,
  Rates of motorcycle helmet use and reasons for non-use among adults and
  children in luang prabang, lao people’s democratic republic, BMC public
  health 15~(1) (2015) 970.

\bibitem{ledesma2015academic}
R.~D. Ledesma, S.~S. L{\'o}pez, J.~Tosi, F.~M. Po{\'o}, Motorcycle helmet use
  in mar del plata, argentina: prevalence and associated factors, International
  journal of injury control and safety promotion 22~(2) (2015) 172--176.

\bibitem{karuppanagounder2016academic}
K.~Karuppanagounder, A.~V. Vijayan, Motorcycle helmet use in calicut, india:
  User behaviors, attitudes, and perceptions, Traffic injury prevention 17~(3)
  (2016) 292--296.

\bibitem{xuequn2011time}
Y.~Xuequn, L.~Ke, R.~Ivers, W.~Du, T.~Senserrick, Prevalence rates of helmet
  use among motorcycle riders in a developed region in {C}hina, Accident
  Analysis \& Prevention 43~(1) (2011) 214--219.

\bibitem{oxley2018academic}
J.~Oxley, S.~O'Hern, A.~Jamaludin, An observational study of restraint and
  helmet wearing behaviour in malaysia, Transportation research part F: traffic
  psychology and behaviour 56 (2018) 176--184.

\bibitem{eby2011naturalistic}
D.~W. Eby, Naturalistic observational field techniques for traffic psychology
  research, in: Handbook of traffic psychology, Elsevier, 2011, pp. 61--72.

\bibitem{dalal2005histograms}
N.~Dalal, B.~Triggs, Histograms of oriented gradients for human detection, in:
  IEEE conference on Computer vision and Pattern Recognition (CVPR), Vol.~1,
  IEEE, 2005, pp. 886--893.

\bibitem{dahiya2016automatic}
K.~Dahiya, D.~Singh, C.~K. Mohan, Automatic detection of bike-riders without
  helmet using surveillance videos in real-time, in: International Joint
  Conference on Neural Networks (IJCNN), IEEE, 2016, pp. 3046--3051.

\bibitem{vishnu2017detection}
C.~Vishnu, D.~Singh, C.~K. Mohan, S.~Babu, Detection of motorcyclists without
  helmet in videos using convolutional neural network, in: International Joint
  Conference on Neural Networks (IJCNN), IEEE, 2017, pp. 3036--3041.

\bibitem{he2016deep}
K.~He, X.~Zhang, S.~Ren, J.~Sun, Deep residual learning for image recognition,
  in: IEEE Conference on Computer Vision and Pattern Recognition (CVPR), IEEE,
  2016, pp. 770--778.

\bibitem{simonyan2014very}
K.~Simonyan, A.~Zisserman, Very deep convolutional networks for large-scale
  image recognition, arXiv preprint arXiv:1409.1556.

\bibitem{szegedy2016rethinking}
C.~Szegedy, V.~Vanhoucke, S.~Ioffe, J.~Shlens, Z.~Wojna, Rethinking the
  inception architecture for computer vision, in: IEEE Conference on Computer
  Vision and Pattern Recognition (CVPR), IEEE, 2016, pp. 2818--2826.

\bibitem{lin2018focal}
T.-Y. Lin, P.~Goyal, R.~Girshick, K.~He, P.~Doll{\'a}r, Focal loss for dense
  object detection, IEEE Transactions on Pattern Analysis and Machine
  Intelligence.

\bibitem{he2017mask}
K.~He, G.~Gkioxari, P.~Doll{\'a}r, R.~Girshick, Mask {R-CNN}, in: IEEE
  International Conference on Computer Vision (ICCV), IEEE, 2017, pp.
  2980--2988.

\bibitem{pigou2018beyond}
L.~Pigou, A.~Van Den~Oord, S.~Dieleman, M.~Van~Herreweghe, J.~Dambre, Beyond
  temporal pooling: Recurrence and temporal convolutions for gesture
  recognition in video, International Journal of Computer Vision 126~(2-4)
  (2018) 430--439.

\bibitem{donahue2015long}
J.~Donahue, L.~Anne~Hendricks, S.~Guadarrama, M.~Rohrbach, S.~Venugopalan,
  K.~Saenko, T.~Darrell, Long-term recurrent convolutional networks for visual
  recognition and description, in: IEEE Conference on Computer Vision and
  Pattern Recognition (CVPR), IEEE, 2015, pp. 2625--2634.

\bibitem{world2017powered}
{World Health Organization}, Powered two-and three-wheeler safety: a road
  safety manual for decision-makers and practitioners, World Health
  Organization, 2017.

\bibitem{Wegman2017road}
F.~Wegman, B.~Watson, S.~V. Wong, S.~Job, M.~Segui-Gomez, Road Safety in
  Myanmar. Recommendations of an Expert Mission invited by the Government of
  Myanmar and supported by the Suu Foundation., Paris, FIA, 2017.

\bibitem{redmon2017yolo9000}
J.~Redmon, A.~Farhadi, {YOLO}9000: Better, faster, stronger, in: IEEE
  Conference on Computer Vision and Pattern Recognition (CVPR), IEEE, 2017, pp.
  6517--6525.

\bibitem{Shen:EECS-2016-193}
A.~Shen,
  \href{http://www2.eecs.berkeley.edu/Pubs/TechRpts/2016/EECS-2016-193.html}{Beaverdam:
  Video annotation tool for computer vision training labels}, Master's thesis,
  EECS Department, University of California, Berkeley (Dec 2016).
\newline\urlprefix\url{http://www2.eecs.berkeley.edu/Pubs/TechRpts/2016/EECS-2016-193.html}

\bibitem{russell2008labelme}
B.~C. Russell, A.~Torralba, K.~P. Murphy, W.~T. Freeman, Label{M}e: a database
  and web-based tool for image annotation, International Journal of Computer
  Vision 77~(1-3) (2008) 157--173.

\bibitem{vondrick2013efficiently}
C.~Vondrick, D.~Patterson, D.~Ramanan, Efficiently scaling up crowdsourced
  video annotation, International Journal of Computer Vision 101~(1) (2013)
  184--204.

\bibitem{girshick2015fast}
R.~Girshick, Fast {R-CNN}, in: IEEE International Conference on Computer Vision
  (ICCV), IEEE, 2015, pp. 1440--1448.

\bibitem{deng2009imagenet}
J.~Deng, W.~Dong, R.~Socher, L.-J. Li, K.~Li, L.~Fei-Fei, Image{N}et: A
  large-scale hierarchical image database, in: IEEE Conference on Computer
  Vision and Pattern Recognition (CVPR), IEEE, 2009, pp. 248--255.

\bibitem{kingma2014adam}
D.~P. Kingma, J.~Ba, Adam: A method for stochastic optimization, arXiv preprint
  arXiv:1412.6980.

\bibitem{everingham2010pascal}
M.~Everingham, L.~Van~Gool, C.~K. Williams, J.~Winn, A.~Zisserman, The pascal
  visual object classes (voc) challenge, International Journal of Computer
  Vision 88~(2) (2010) 303--338.

\bibitem{salton1983introduction}
G.~Salton, M.~J. McGill, Introduction to modern information retrieval, McGraw
  Hill Book Co., 1983.

\bibitem{chollet2015keras}
F.~Chollet, et~al., Keras, \url{https://keras.io} (2015).

\bibitem{hyder2019measurement}
A.~A. Hyder, Measurement is not enough for global road safety: implementation
  is key, The Lancet Public Health 4~(1) (2019) e12--e13.

\bibitem{world2006helmets}
{World Health Organization}, Helmets: a road safety manual for decision-makers
  and practitioners, Geneva: World Health Organization, 2006.

\end{thebibliography}

\end{document}